\pdfoutput=1

\documentclass[11pt]{article}

\usepackage[]{acl}

\usepackage{graphicx}

\usepackage{times}
\usepackage{latexsym}

\usepackage[T1]{fontenc}

\usepackage[utf8]{inputenc}

\usepackage{microtype}

\usepackage{inconsolata}

\usepackage{booktabs}

\title{Describing Images \textit{Fast and Slow}: Quantifying and Predicting the Variation in Human Signals during Visuo-Linguistic Processes}

\author{Ece Takmaz \and Sandro Pezzelle \and Raquel Fern\'andez\\
	Institute for Logic, Language and Computation\\University of Amsterdam\\
	\texttt{\{ece.takmaz|s.pezzelle|raquel.fernandez\}@uva.nl} \\}

\begin{document}
\maketitle
\begin{abstract}

There is an intricate relation between the properties of an image and how humans behave while describing the image. This behavior shows ample variation, as manifested in human signals such as eye movements and when humans start to describe the image. Despite the value of such signals of visuo-linguistic variation, they are virtually disregarded in the training of current pretrained models, which motivates further investigation. Using a corpus of Dutch image descriptions with concurrently collected eye-tracking data, we explore the nature of the variation in visuo-linguistic signals, and find that they correlate with each other. Given this result, we hypothesize that variation stems partly from the properties of the images, and explore whether image representations encoded by pretrained vision encoders can capture such variation. Our results indicate that pretrained models do so to a weak-to-moderate degree, %
suggesting that the models lack biases about what makes a stimulus complex for humans and what leads to variations in human outputs.

\end{abstract}

\section{Introduction}

Humans can capture the gist of an image usually incredibly fast  -- 100 msec could be enough \citep{oliva2005gist, oliva2006building}; however, they would need more time to act on an image. %
For instance, human behavior while describing images illustrates the intricacies of visuo-linguistic processes. 
There may be repetitions, silent intervals and disfluencies, with considerable degrees of variation in what is uttered across speakers. The period prior to %
the utterance involves perceiving the image, conceptualizing the message, retrieving the labels of the entities %
to mention, formulating and preparing to articulate a grammatical and relevant utterance \citep{levelt, Slobin2003-SLOLAT}. %

\begin{figure}[h]
    \centering
    \begin{minipage}[h]{0.45\columnwidth}
    \includegraphics[width=\columnwidth]{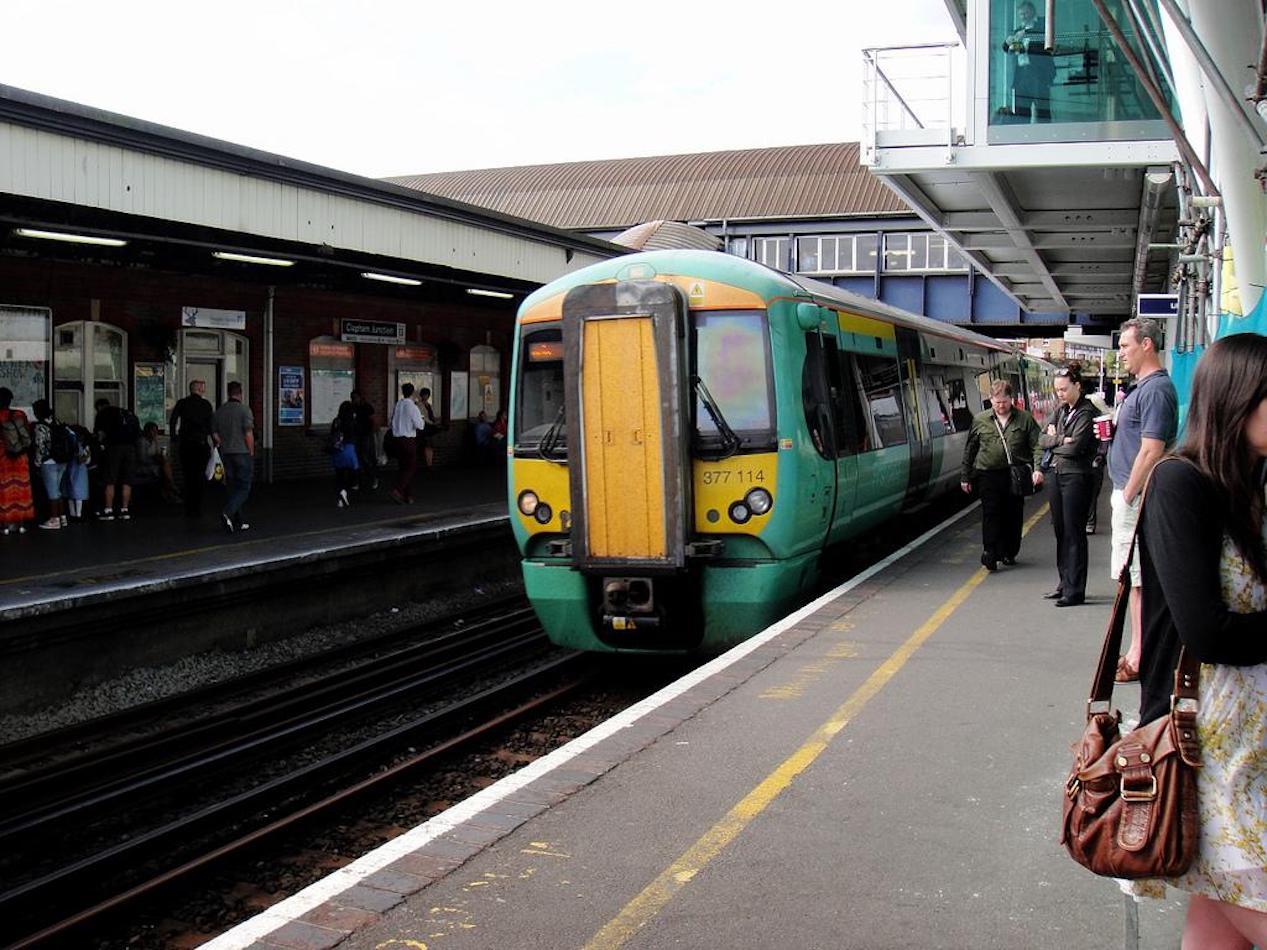}
    \vspace{-0.8cm}
    \caption*{Min: 1.69 sec}
  \end{minipage}
  \hfill
  \begin{minipage}[h]{0.45\columnwidth}
    \includegraphics[width=\columnwidth]{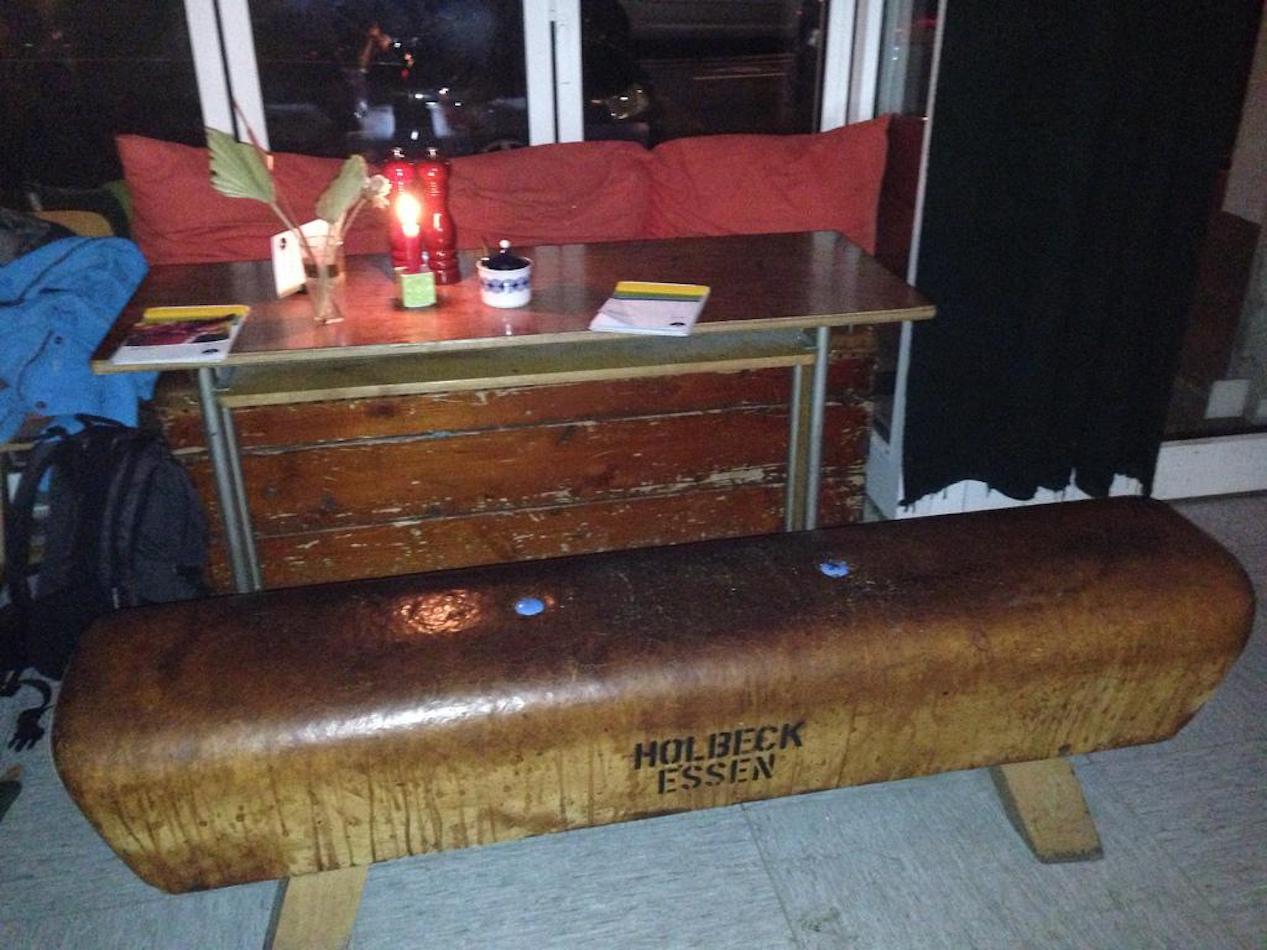}
    \vspace{-0.8cm}
    \caption*{Max: 7.07 sec}
  \end{minipage}
    \caption{%
    The images with the minimum and maximum mean speech onsets across speakers in the dataset. The image with the maximum onset also elicits the highest variation in the first nouns of the descriptions.} %
    \label{fig:exp_onset}
\end{figure}
As a result, we observe variations in \textbf{speech onsets}, as in Figure~\ref{fig:exp_onset}, which could be indicative of the relative cognitive complexity induced by the images~\citep{integrating, gatt}. In addition, different speakers might start their utterances with different words (\textbf{starting points}, see~\citealp{starting}), continuing to produce a varied set of image descriptions (\textbf{linguistic variation}) with \textbf{variation in gaze}. These signify the intricate cross-modal relation between visual and linguistic processes in humans~\citep{griffinbock, linear}. %

Although human data can be rich in behavioral signals, current pretrained multimodal models virtually never receive information about such signals during training. The models generate descriptions without necessarily modeling how human processes unfold. %
For instance, deep neural networks can output words at the same rate even for images that would result in diverse speech behavior by humans due to complexity or ambiguity. Moreover, there is a gap between the manner in which humans perceive stimuli as compared to how large models process them. Model-predicted surprisal values for linguistic input can be lower than human surprisal, possibly due to the massive size of the training data and the number of model parameters~\citep{vanschijndellinzen, arehalli-etal-2022-syntactic, oh2023transformerbased, 10.1162/tacl_a_00548}. Models also display different patterns of visual attention compared to humans~\citep{das-etal-2016-human}. %

We argue that it is essential to consider human signals %
such as speech onsets and looking times, as they reflect the complexity and ambiguity of visuo-linguistic tasks~\citep{integrating, gatt, Meulen2001EyeMD, Meyer2000PhonologicalPE, miltenburg2018DIDEC}. It is therefore desirable if models encode what leads to variations in such signals %
to help generate image descriptions in a way that is aligned with human processing and with types of variation observed in human data~\citep{van-miltenburg-etal-2018-measuring}. To this end, several applications have exploited human gaze to enhance image captioning and visual question answering models~\citep{sugano2016seeing, He2019HumanAI, takmaz-etal-2020-generating, sood-etal-2021-vqa, Sood_2023_CVPR}. Still, the relation between gaze on images and language is not widely researched in NLP~\citep{alacam-etal-2022-modeling-updated}.

We first explore the natural dynamics in visuo-linguistic processes using The Dutch Image Description and Eye-tracking Corpus~\cite[DIDEC;][]{miltenburg2018DIDEC}. This corpus provides gaze and speech data concurrently collected while participants describe images depicting real-life scenes. We preprocess the DIDEC dataset extensively, and propose metrics to quantify the variation in visual and linguistic modalities. We reveal for the first time significant correlations between speech onsets, variation in starting points, descriptions and gaze. 

We hypothesize that this variation is partly due to the properties of the images, and that similar images would elicit similar amounts of variation. Given the superior performance of pretrained encoders that are widely used in multimodal models, we investigate whether visual encoders such as CLIP~\cite{clip} and ViT~\cite{dosovitskiy2021an} capture information regarding the variation in visuo-linguistic signals.\footnote{Code and data available at \url{https://github.com/ecekt/visuolinguistic_signal_variation}.} This is akin to probing pretrained models for meaningful syntactic and semantic information; see~\citealp{conneau-etal-2018-cram}. Using a similarity-based prediction method~\citep{ANDERSON201644}, we find that the pretrained encoders capture variation in signals to a limited extent. Our findings suggest that underlying factors leading to variation are encoded rather weakly by pretrained models. With our work, we aim to direct attention towards the importance of the information contained in such signals and the variation thereof when crowdsourcing data as well as during model development.

\section{Background}

We first give an overview of visuo-linguistic processes in humans in Section~\ref{multimodal}, and then, in models in Section~\ref{mmnlp}.

\subsection{Visuo-Linguistic Processes in Humans}
\label{multimodal}

\paragraph{Cross-modal processes} Describing images requires the linear unfolding of complex cross-modal processes between vision and language \citep{henderson2013interface, griffinbock, gleitman2007give, scanpatterns, linear, HENDERSON201715}. There exist several theories regarding how the `linearization'~\citep{levelt} takes place in sentence formulation %
in relation to visual processes~\citep{whylook, meyer,linear}. These theories consider the speaker's knowledge %
and expectation regarding the contents of the image, as factors affecting the allocation of gaze and the formulation of a description~\citep{HENDERSON201715, linear}. %
In addition, the way people look at an image changes based on the task at hand \citep{yarbus1967eye, buswell1935people, castelhano}, with similar sequences of fixations (\textit{scanpaths}) leading to the production of similar sentences~\citep{scanpatterns}. Therefore, we hypothesize that the variation in language production and eye movements could be correlated.  %

\paragraph{Starting points} %
A sentence must have a starting point, given that words need to be uttered in a linear order~\citep{levelt}. We take the first uttered noun as the starting point of image descriptions. The focus on nouns is motivated by the fact that gaze scanpaths are frequently represented by sequences of object categories, which tend to be expressed by nouns. Additionally, the order of mention of these categories is the point of interest in linearization studies that investigate language production parallel to visual processes~\citep{linear}. 
Starting points can be selected based on a variety of factors (canonical word order of the language, perspective of the speaker, complexity of the planned sentence; see~\citealp{starting}). When describing images, visual properties of an image influence how a sentence begins and unfolds~\citep{putting}.
These findings signify how the selection of starting points can be influenced by a set of complex visuo-linguistic factors.  %

\paragraph{Variation in image descriptions} 
People generally describe images with some variation. %
\citet{Jas_2015_CVPR} report that images with people and large objects tend to be described more specifically, whereas generic buildings, ambiguous scenes and images with less-important objects tend to elicit more varied descriptions. The degree to which the descriptions of an image vary is referred to as `image specificity' by \citet{Jas_2015_CVPR}, who propose an automatic metric to quantify it using the similarity scores between the WordNet paths of words in descriptions ~\citep{miller-1994-wordnet}. \citet{miltenburg2018DIDEC} explore image specificity in the corpus that we use in this study, utilizing word2vec vectors~\citep{mikolov2013efficient} to compute the similarity scores. %
They find that the variation in descriptions is only to a limited extent due to the image's contents as there also seems to be an effect of language (English vs. Dutch). Additionally, their results indicate that attention maps extracted using gaze data do not help predict image specificity~\citep{miltenburg2018DIDEC}. In this work, we also quantify and predict image specificity proposing different approaches. %

\paragraph{Speech onsets} Slower speech onsets indicate that a deliberate, effortful process is taking place, as compared to fast onsets; as claimed in the dual process theory~\citep{WASON1974141, kahneman}. Various intertwined linguistic and visual processes  modulate speech onsets and the latency of referring to an object %
~\citep{Meyer2000PhonologicalPE, integrating}, %
such as the contents of an image and the locations of the objects~\citep{gatt, esaulova2019DescribingEC}. %
This indicates that speech onsets are strongly linked to image features. Given the importance of speech onsets in relation to visuo-linguistic processes and the cognitive requirements of a task, the mean speech onset induced by an image across speakers is one of the signals we focus on.

\subsection{Multimodal NLP}
\label{mmnlp}

\paragraph{Pretrained models} Many recent multimodal models employ frozen pretrained unimodal models and combine them with either no further training or via trained lightweight mapping networks~\citep{lens, flamingo, mapl, tsimpoukelli2021multimodal, li2023blip2, mokady2021clipcap, visualgpt}. Particularly, the visual encoder of the CLIP model~\citep{clip} has been utilized in these models as a foundation model with strong zero-shot capabilities that improves multimodal models~\citep{shen2022how}. 

By training classifiers on top of visual encoders, \citet{constraints} predict the existence of linguistic features such as passive voice and the use of numeral expressions in image descriptions, and indicate that the selection of such linguistic features is constrained by visual features. These findings point to the underlying capabilities of pretrained models pertaining to human cognitive processes.

\paragraph{Human signals in NLP}

Most previous research into the use of human signals focuses on text-only cases~\citep{klerke-etal-2016-improving, barrett-etal-2018-sequence, barrett-etal-2016-weakly, mishra, hollenstein-etal-2021-cmcl, hollenstein-etal-2022-cmcl, hollenstein-etal-2021-multilingual, pouw-etal-2023-cross,cogbert, cogalign, gazby, synthesizing, ijcai2020p0683, ijcai2020p689}. However, the relationship between human gaze on images and language production, and its potential contribution to computer vision and NLP has been investigated even before the existence of pretrained models~\citep{6618945}. Research into whether the attention distributions in multimodal models correlate with human attention reveals contrasting findings~\citep{das-etal-2016-human, gella-keller-2018-evaluation, He2019HumanAI, sood-etal-2021-vqa}. %
Several works show that the use of human gaze enhances image captioning and visual question answering~\citep{sugano2016seeing, He2019HumanAI, takmaz-etal-2020-generating, sood-etal-2021-vqa, Sood_2023_CVPR}. Yet, modeling gaze in conjunction with linguistic processes is still an under-explored area in NLP~\citep{alacam-etal-2022-modeling-updated}.

In our work, we investigate the variation of a set of human signals in a corpus, as well as whether pretrained vision encoders can encode information related to these signals. Although such models are shown to be very effective in multimodal tasks, they are still under-explored from this point of view.

\section{Data}
We aim to explore the variation in human signals in visuo-linguistic processes and whether pretrained models can capture such variation in a realistic setup. A dataset consisting of simultaneous language production and eye movements over complex images would enable such an exploration. 
Therefore, we opt for using the DIDEC corpus~\citep{miltenburg2018DIDEC} instead of other existing image description datasets with eye-tracking, as this corpus allows us to delve into the dynamics of visual and linguistic processes in parallel. There exist few datasets containing such information, which we did not opt for utilizing, as they differ in their tasks \citep[narratives;][]{vaidyanathan-etal-2018-snag}, or the processing steps the authors have taken, e.g., only a small subset of the captions were checked manually~\citep{vaidyanathan-etal-2018-snag}; the authors sample one gaze point every 4 points~\citep{He2019HumanAI}. The DIDEC dataset comprises manually checked descriptions of high quality, and the gaze data is provided in a raw format enabling custom processing. %

We use the `production viewing' subset of DIDEC, which contains spoken descriptions for 307 real-life images originating from the MS COCO dataset \cite{coco}, with high-quality eye-tracking data.\footnote{The other subset contains data from `free viewing', where the participants simply looked at the images for 3 seconds.} 45 participants describe $\approx$ 102 images without a time limit. On average, each image has 15 descriptions (4604 in total). Next, we explain how we extract features corresponding to human signals in visuo-linguistic processes from this dataset, to obtain 4586 descriptions with speech onsets, starting points, and fixated regions.

\subsection{Visual Data}
Using the raw gaze samples in DIDEC~\citep{miltenburg2018DIDEC} labeled as fixations, saccades, and blinks, we create fixation windows by treating saccades and blinks as boundaries~\citep{salvucci}. The gaze samples in the fixation window are then put into a list, skipping the ones that fall outside the boundaries of the images. To visually represent a fixation, we feed its gaze points as coordinate prompts to the Segment Anything Model \citep[SAM;][]{segm}. Using the prompts, this model predicts the objects the gaze corresponds to, and outputs masks corresponding to fixated regions. We use the ViT-L version of the model building on vision transformers~\citep{dosovitskiy2021an}, as it achieves good performance~\citep{segm}. 
We obtain a single mask per fixation window. The masks sometimes span non-contiguous regions; therefore, we utilize the bounding box based on the $x$-$y$ limits of the predicted mask. %

\subsection{Linguistic Data} 
\label{descr}
\paragraph{Speech onsets}
The dataset supplies audio files for spoken descriptions and their transcripts. To extract word-level timestamps, we use WhisperX~\citep{bain2022whisperx} based on Whisper~\citep{radford2022robust}.\footnote{The model for obtaining alignments for audio in Dutch: \texttt{jonatasgrosman/wav2vec2-large-xlsr-53-dutch}} 
We relay the transcripts directly into the alignment function of WhisperX. The output contains the start and end timestamps of each word. This also allows us to extract information regarding when the participants start talking, i.e., speech onsets. %
The mean speech onset is 3.42 sec, and the median is 2.65 sec. We observe variation across participants and images, as the onsets can go up to 25.37 sec with a standard deviation (SD) of 2.45.

\begin{figure*}[]
    \centering
    \includegraphics[width=\textwidth]{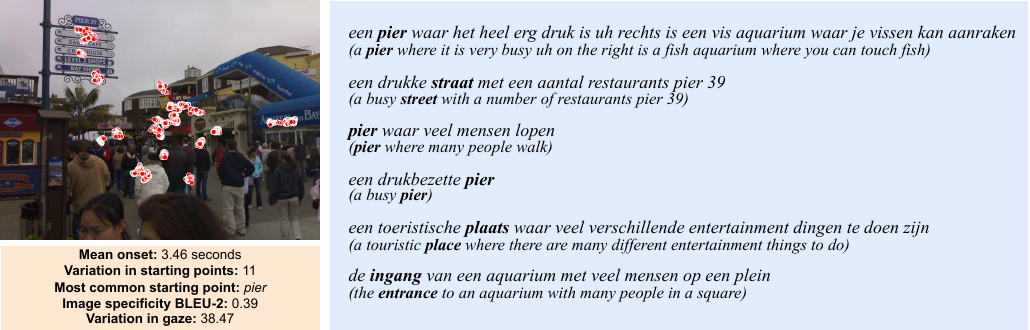}
    \caption{An image with its variation scores, a subset of its descriptions (along with the English translations in parentheses), and the eye movements of a single participant. In the descriptions, the words in boldface indicate the starting points in Dutch and their equivalents in English.}
    \label{fig:exscores}
\end{figure*}

\paragraph{Starting points}
We use the spaCy library for tokenization, part-of-speech tagging, and lemmatization of the words in the descriptions.\footnote{\texttt{nl\_core\_news\_lg} pipeline from \url{https://spacy.io/}} For Dutch, the library provides 3 models (small, medium, and large). Upon manual inspection of 50 random samples from the data processed by each model, we opted for the large model, which yields the least number of errors. See Appendix~\ref{spacy} for more details.

\section{Variation in Human Signals}
\label{variation}
We first delve into the nature of the variation across humans per image in the DIDEC dataset. Our focus is on uncovering potential correlations between the variations in human signals in visuo-linguistic processes. We first explain how we quantify each signal and its variation, see Figure~\ref{fig:exscores} for an example image with all of its variation scores. Then, we conduct pairwise correlation analyses between the 4 variables. If there exist correlations between variations across signals, one can speculate that at least part of the correlation stems from %
the image, with the rest being potentially due to factors such as viewing order, priming and cognitive load.  %

\subsection{Variation in Speech Onsets}
We inspect the mean and SD of speech onsets per image, see histograms in Appendix~\ref{distonset}. The mean onsets per image range between 1.69 and 7.07 seconds, constituting a non-normal distribution skewed towards shorter onsets %
($p < .001$, 65.77\% of the onsets shorter than the mean onset). For some images, some participants start talking immediately; whereas, in other cases, they wait for a considerable amount of time before speaking. This observation resonates with the fast and slow systems from the dual process theory~\citep{WASON1974141, kahneman}, suggesting that more complex processes are recruited while describing certain images. %
However, even for a single image, the participants might start speaking at varying times (with SD per image ranging from 0.44 to 6.33). This suggests that various factors are at play while describing images, such as contextual and speaker-specific effects. 

To have a better picture of onset variation, we compare the onsets for an image against each other. Leaving one onset out of the set of onsets for an image, we calculate the average of the rest ($\approx 14$ onsets). The difference between the average and the left-out onset corresponds to error. We perform this calculation for each sample. Then, we take the mean over all the samples, which yields an error of 1.625 seconds. %
This error is a proxy for the average variation over the participants, which suggests that there is a difference in response times across humans when prompted with the same image.

The DIDEC corpus comes with 3 mutually-exclusive image subsets called `lists'. Each participant views only one list. We find that the mean onsets in List $2$ are significantly shorter than the other two sets ($p < .001$, independent samples t-test). Since both the images and the participants are different across lists, it is not straightforward to separate their effects. See Appendix~\ref{ppncorr} for a participant-based analysis of mean onsets.
 
\subsection{Variation in Starting Points}
Counting the first nouns of image descriptions reveals that there is an imbalance in the starting points in the data. The participants utter words such as \textit{man}, \textit{person}, \textit{woman}, \textit{bus} and \textit{street} most frequently as the first noun of a description (370, 238, 221, 174, 141, respectively, constituting in total 25\% of the samples). This is potentially due to the salience of such entities and their frequency in the images. We represent the variation in starting points by the number of unique starting points uttered per image, yielding $mean = 6.45$, $min = 1$, $max = 13$. These values indicate that some images elicit the same first nouns, whereas some others prompt the production of a range of starting points.\footnote{We compute variation in starting points with respect to exact noun lemma matches, without considering synonyms that could refer to the same object. We believe that this captures the type of variation that is of interest for starting points, since lexical choices reflect categorization and conceptualization of objects that can be affected by the visual context in which the object is situated \citep{GUALDONI2023104459}.}  %

\subsection{Variation in Full Descriptions}
\label{imgspec_onset}
Each image can be described in distinct ways, both in terms of the words uttered and their order. We quantify the \textit{linguistic variation} in image descriptions, following a different approach compared to \citet{Jas_2015_CVPR} and \citet{miltenburg2018DIDEC}. We adopt a widely used natural language generation metric, BLEU~\citep{Papineni:2002}. This metric computes n-gram-based precision scores between a generated sentence and a set of references. We opt for the bigram version (BLEU-2), since we are mostly interested in the surface form variation of words, and to a limited extent, the sequences of words. BLEU-2 allows us to measure the linguistic variation in descriptions independently of a pretrained model.\footnote{See Appendices~\ref{bertje} and ~\ref{metrics} for a semantic variation metric we propose using Dutch BERT-based representations~\cite[BERTje;][]{devries2019bertje}, another combining BERTje and BLEU-2-based variation, as well as a comparison to human annotations provided by \citet{Jas_2015_CVPR}.} We calculate the BLEU-2 score between a description and the remaining descriptions for the image constituting the reference set. Then, we take the average over all descriptions of an image.\footnote{This metric is similar to Self-BLEU~\citep{selfbleu}, which was proposed to calculate the diversity of the sentences generated by a model. In Self-BLEU, each generated sentence is compared to the rest of the generated sentences within a document, and an average of the whole set is computed to indicate how varied a model’s generations are.
} This method yields an extensive range of normally distributed scores ($\mu = 0.53, min = 0.25, max = 0.81$).

\subsection{Variation in Gaze}

The variation in eye movements has been quantified in various ways in the literature: scanpath complexity, dispersion of the heatmap of gaze on an image, entropy of the gaze distribution~\citep{integrating}. We propose a distance metric based around the contents of fixated regions and their orders. We represent a scanpath in the form of a sequence of fixation bounding boxes represented as $(x_1, y_1, x_2, y_2)$. Given two scanpaths $S_1$ and $S_2$, for each fixation box in $S_1$, we find the most similar box in $S_2$ that yields the highest ratio of intersection over union (IoU) between the bounding boxes. The IoU dissimilarity $(1-IoU)$ as well as the normalized positional distance %
between these boxes are summed up. This step is performed for all fixation boxes in $S_1$. The total gives us a comparison score for two scanpaths. We compare $S_1$ to all the other scanpaths for the same image and then, take the average. Each scanpath for the image is compared to the rest of the related scanpaths in the same way. This yields 15 image-scanpath variation scores, whose mean corresponds to the gaze variation score of a single image. The higher this score is, the more variation exists in the gaze modality. We obtain a range of gaze variation scores for the whole set ($mean = 24.00$, $min = 11.22$, $max = 38.79$).

\subsection{Correlation between Variations}
\label{corrvar}
\begin{figure}
    \centering
    \hspace{-0.8cm}
    \includegraphics[width=0.8\columnwidth]{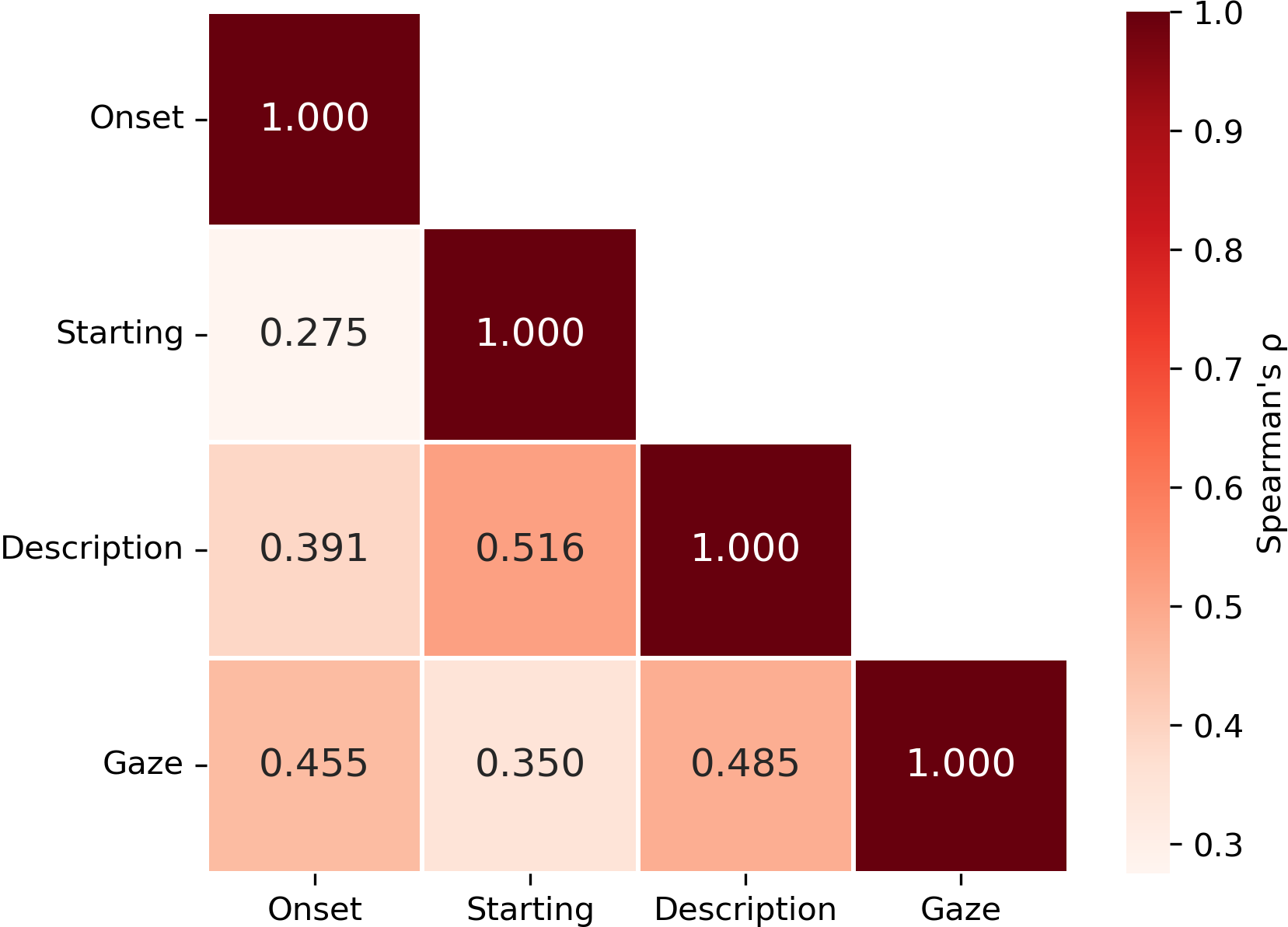}
    \caption{Spearman's correlation coefficients between the mean onsets per image (Onset), the variation in starting points (Starting), BLEU-2-based variation in full descriptions (Description), and the variation in gaze (Gaze) in the full dataset. Since higher BLEU scores mean less variation unlike the trends in the other measures, we utilize $1-BLEU$ for better interpretability. All of the correlations are significant, $p < .001$.}
    \label{fig:corrmatrix}
\end{figure}

In the previous subsections, we have quantified the variation in speech onsets, starting points, descriptions and gaze per image, see Appendix~\ref{app_examples} for the images with the minimum and maximum scores across our variables of interest. We now turn to the correlation between the variation types. Since the initial common point is the image itself, we hypothesize that image features contribute to varying levels of variation in different modalities. We run Spearman's correlation between each type of variation.\footnote{We conduct Spearman’s rank correlation analysis to uncover monotonic relations in the data. This type of correlation does not assume a particular distribution of the data (non-parametric, as opposed to Pearson’s normality assumption). Since some of the signals we have investigated are non-normally distributed (e.g., speech onsets), and the dataset is relatively small, we opted for Spearman.} When interpreting the magnitudes of the correlation coefficients, we use the terminology suggested by \citet{Prion2014MakingSO}. See Figure~\ref{fig:corrmatrix} for all correlation results. 

We find a significant negative correlation, approaching moderate effect, between BLEU-2-based linguistic variation and the mean onset of an image (Spearman's $\rho = -0.391, p < .001$, see Appendix~\ref{corrline} for the regression line). This means that speakers start describing images that yield more similar descriptions earlier.\footnote{Unlike this correlation, we find that speech onsets are not correlated with how many words or nouns are uttered.}
In addition, as starting points vary, image descriptions  become less similar (moderate, Spearman's $\rho = -0.516, p < .001$), indicating that initial deviations continue until the end of language production. 

We find that the variation in gaze significantly correlates with speech onsets (moderate, Spearman's $\rho = 0.455, p < .001$); the variation in starting points (weak, Spearman's $\rho = 0.350, p < .001$); and the variation in full descriptions (moderate, Spearman's $\rho = -0.485, p < .001$). These outcomes indicate that high variation in gaze tends to co-occur with longer onsets, high variation in starting points, and less similarity in descriptions.\footnote{Investigating the correlation between these types of variation and the number of objects in an image is not straightforward, as current object detection algorithms annotate images exhaustively, yielding a high number for many images.}

The correlations reveal a connection between the variation in visual and linguistic modalities. %
We hypothesize that the underlying reasons for such variation partly reside in the features of an image, echoing the claims by \citet{Jas_2015_CVPR} and \citet{constraints}. In this sense, similar images are expected to elicit similar amounts of variation. Hence, the results motivate our research into whether image features as encoded by pretrained models can capture the variation in gaze and language.

\section{Similarity-based Prediction}
\label{sec:notrain}
In light of the correlation findings in Section~\ref{variation}, we expect image features to be predictive of the variation in visuo-linguistic signals to some extent. We explore if the similarity scores between image features encoded by pretrained models would be meaningful when capturing variation in human signals. In particular, we hypothesize that the signals that are more internal to the pretrained models' training objectives would be captured better. For instance, CLIP was trained with respect to an image-to-text alignment objective~\citep{clip}; hence, it would be reasonable to expect that signals that are more inherent to the visual and language data could be encoded better compared to speech onsets, which are never seen by the model.

\paragraph{Approach} We employ an approach that was proposed as an alternative to training regression models and representational similarity analysis, for predicting fMRI signals given linguistic input \citep{ANDERSON201644}. Using the similarities between model-encoded stimuli (embeddings of concepts) and the corresponding fMRI responses, the authors predict the fMRI signals for novel stimuli for which embeddings exist. This approach has been utilized to assess the extent to which deep neural networks capture brain representations in language-only and visually grounded setups~\citep{anderson-etal-2017-visually, https://doi.org/10.1111/cogs.13388, Bruera2023.08.23.554436}. We explain how we operationalize this extrapolation method for our purposes in Section~\ref{notrain_imgspec}. As this approach does not require training, it is suitable for shedding light on the predictive power of pretrained image representations, given the small size of the dataset we use. We determine the splits based on the images. Hence, to mitigate imbalance issues, we create 50 random split setups with $\sim$90\% training (277 images) and $\sim$10\% test sets (30 images), and report results on the average of these 50 setups. Across setups, the training sets have similar representative powers in terms of their CLIP vector similarities to the images in the corresponding test sets. %

\paragraph{Visual encoders}
To encode the images, we exploit three visual encoders: CLIP, ViT, and a randomly initialized CLIP model (without training at all). We use the ViT-B/32 version of CLIP's visual encoder~\citep{clip}, and extract the final 512-dimensional output for each image. Since this encoder has been trained in coordination with CLIP's textual encoder~\citep{clip}, we expect it to capture not only vision-related features, but also properties that are aligned with language. In addition, we test the representations of a purely visual encoder trained on object recognition, ViT~\citep{dosovitskiy2021an}. We extract the last hidden states from ViT, and use the vector corresponding to the \texttt{[CLS]} token as the image representation. %
Finally, we also experiment with a randomly-initialized version of CLIP (\textsc{RndCLIP}), along the lines of what \citet{constraints} did to avoid the information learned during pretraining.

\subsection{Predicting the Variation in Descriptions}
\label{notrain_imgspec}
From the training set, we retrieve $k$ images that are closest to the target image---the image for which we predict a signal variation score---based on their representational similarities, echoing the k-nearest neighbors algorithm. The final score is the weighted average of the variation found in the neighboring images. The weights correspond to the similarity scores between the retrieved images and the target image. %

As depicted in Table~\ref{tab:spec}, we find significant, yet weak, positive correlations for almost half of the 50 split configurations both for CLIP and ViT, with no meaningful correlations for \textsc{RndCLIP}. %
CLIP slightly outperforms ViT, suggesting that language alignment in the visual modality yields a potential benefit in estimating the variation in descriptions.

\begin{table}[h]
\centering
\resizebox{0.9\columnwidth}{!}{
\begin{tabular}{lccc}
\toprule
\textbf{Model}        & \textbf{Coefficient} & \textbf{Sig.} & \textbf{Loss} \\ \midrule %
\textbf{CLIP}    &     \textbf{0.3380}        &     \textbf{27}         &  0.0738  \\ %
\textbf{ViT}     &     0.3135        &    23          &  \textbf{0.0723}   \\ %
\textsc{\textbf{RndCLIP}} &     0.0472        &      3        & 0.0744  \\ \bottomrule %
\end{tabular}
}
\caption{Predicting variation in descriptions with the similarity-based approach, $k=277$. Averages over 50 random splits. `Coefficient' and `Sig.' correspond to Spearman's $\rho$ correlation coefficient and how many runs out of 50 yield significant correlations with $p < 0.05$.} %
\label{tab:spec}
\end{table}

The loss corresponds to the average difference between the predicted and target scores across the dataset. The losses are similar across encoder types despite the differences in correlations. Since this method makes predictions based on the ground truth outputs of the retrieved set, it is likely that the predictions remain in a similar range.

\subsection{Predicting Onset}
\label{notrain}
We perform the similarity-based prediction approach outlined in Section~\ref{notrain_imgspec} to predict mean speech onsets per image. Since longer onsets can be associated with more cognitively demanding images, we are interested in the average onset elicited by each image. %
The results (see  Table~\ref{tab:ons}) indicate that, by using a larger sample of CLIP-encoded images, we can obtain predictions weakly correlating with the target onsets. The differences in the results when using different $k$ values suggest that the choice of the retrieval set limits the boundaries of the predictions, even though the median image similarity score for $k=1$ is 0.77 in the dataset.

\begin{table}[h]%
\centering
\resizebox{\columnwidth}{!}{
\begin{tabular}{lcccc}
\toprule
\textbf{Model}        & \textbf{Coefficient} & \textbf{Sig.} & \textbf{Loss} & \textbf{Range} \\ \midrule
\textbf{CLIP-277}   &    \textbf{0.2981}         &        \textbf{18}      & 0.8216    & 3.37 - 3.50 \\ %
\textbf{CLIP-10}     &     0.2500        &     10       & \textbf{0.7989}  &   2.60 - 4.37 \\
\textbf{CLIP-5} &       0.2265       &     14         &  0.8149  & 2.26 - 4.81 \\ %
\textbf{CLIP-1} &      0.0640       &     4         &  1.0746 & 1.69 - 6.39
\\ %
\textbf{ViT}     &      0.2428       &    17          &  0.8072  & 3.11 - 3.67 \\
\textsc{\textbf{RndCLIP}} &      0.0350       &   3           & 0.8249   & 3.38 - 3.47 \\ \bottomrule
\end{tabular}
}
\caption{Predicting mean speech onsets with the similarity-based approach. The numbers in the model names correspond to $k$ when retrieving closest images from the training set. \textsc{RndCLIP} and ViT with $k=277$. `Range' is the range of the predictions for the test set.} %
\label{tab:ons}
\end{table}

When we use 277 images encoded with ViT to obtain the image similarities, the correlation is weaker than the same setup with CLIP. When we encode the images with \textsc{RndCLIP}, although the loss is quite similar to the other setups, there is no meaningful correlation.
The predictions in general center around the mean onset, as they are based on the outputs from the retrieval set.

\subsection{Predicting Starting Points}
We utilize the similarity-based prediction algorithm to predict the first uttered nouns of the descriptions. Since this is a subtask of generating descriptions, we consider this an interesting use case. For each image, we represent the most common first noun as a one-hot vector (with the dimensions being 739, corresponding to the size of the first-noun vocabulary of the whole dataset). We report the accuracy of predicting the correct starting point. %

\begin{table}[h]%
\centering
\resizebox{\columnwidth}{!}{
\begin{tabular}{lcc}
\toprule
\textbf{Model}        &\textbf{ $k=277$} & \textbf{$k=10$}  \\ \midrule
\textbf{CLIP}    &     13.00\%  & \textbf{31.73}\% \\ %
\textbf{ViT}    &     \textbf{26.47}\%  & 30.53\%\\ %
\textsc{\textbf{RndCLIP}} & 11.27\% & 10.40\% \\ %
{\textbf{Baseline - Random}} &  4.00\% & 4.00\% \\ %
{\textbf{Baseline - Most common}} & 11.27\% & 11.27\% \\ \bottomrule
\end{tabular}
}
\caption{Predicting starting points with the similarity-based approach and the baselines, percentage of correctly identified starting points for different $k$ values.}
\label{tab:notrainsp}
\end{table}

As illustrated in Table~\ref{tab:notrainsp}, all setups attain scores that outperform the baseline where we predict random starting points %
(theoretically, for a uniform distribution of starting points, $1/739=0.14\%$). We also predict the most common starting point (`man'), which performs similarly to \textsc{RndCLIP}. %
With pretrained encoders, it is better to utilize lower $k$ to attain better accuracy, since very similar images likely contain similar objects that are mentioned earlier in the utterances. Both CLIP and ViT show similar performances when $k=10$, hinting at the relation between their training objectives and starting points, which often correspond to the most salient entity in the image. %

\subsection{Predicting the Variation in Gaze}
We apply the similarity-based approach to predict the variation in gaze. The results (Table~\ref{tab:gazevar}) reveal that the gaze variation can be approximated to a moderate extent with CLIP. Using a smaller retrieval set is beneficial, suggesting a strong link between image properties and the variation in gaze. Since CLIP has a powerful visual encoder~\citep{shen2022how}, it is reasonable that the similarities between image features encoded by CLIP seem more meaningful when predicting the variation in gaze. %

\begin{table}[h]%
\centering
\resizebox{\columnwidth}{!}{
\begin{tabular}{lcccc}
\toprule
\textbf{Model}        & \textbf{Coefficient} & \textbf{Sig.} & \textbf{Loss} & \textbf{Range} \\ \midrule
\textbf{CLIP-277}   & 0.4035 & 30 & 4.0200 & 23.55 - 24.45 \\ %
\textbf{CLIP-10}     & 0.4253 & 35 & 3.5774 & 17.05 - 29.63\\ %
\textbf{CLIP-5} &  0.4435 & 33 & \textbf{3.5707} &  15.43 - 32.92\\ %
\textbf{CLIP-1} &  \textbf{0.4687}  & \textbf{39} & 3.8889 &11.22 - 38.79\\ %
\textbf{ViT}     & 0.3801  & 28 & 3.8847 & 22.62 - 25.67 \\ %
\textsc{\textbf{RndCLIP}} & 0.0109 & 2 & 4.0571 &  23.76 - 24.26 \\ \bottomrule
\end{tabular}
}
\caption{Predicting gaze variation using the similarity-based approach. Targets range between 11.22 and 38.79.}
\label{tab:gazevar}
\end{table}

The outcomes are in line with our hypothesis that signals that could be considered more internal to the models' training objectives would be captured better, whereas external signals can be captured weakly. For instance, speech onsets and surface form variation in descriptions can be deemed external to CLIP's space. Therefore, we claim that there could be room for incorporating such external signals when training or fine-tuning pretrained multimodal models, and the models would benefit from such signals. It should be noted, though, since human processes are complex, there could be extraneous factors beyond image features that influence variation, which makes it difficult for models to capture these signals perfectly.

\begin{figure}[h]
    \centering
    \begin{minipage}[h]{0.44\columnwidth}
    \includegraphics[width=\columnwidth]{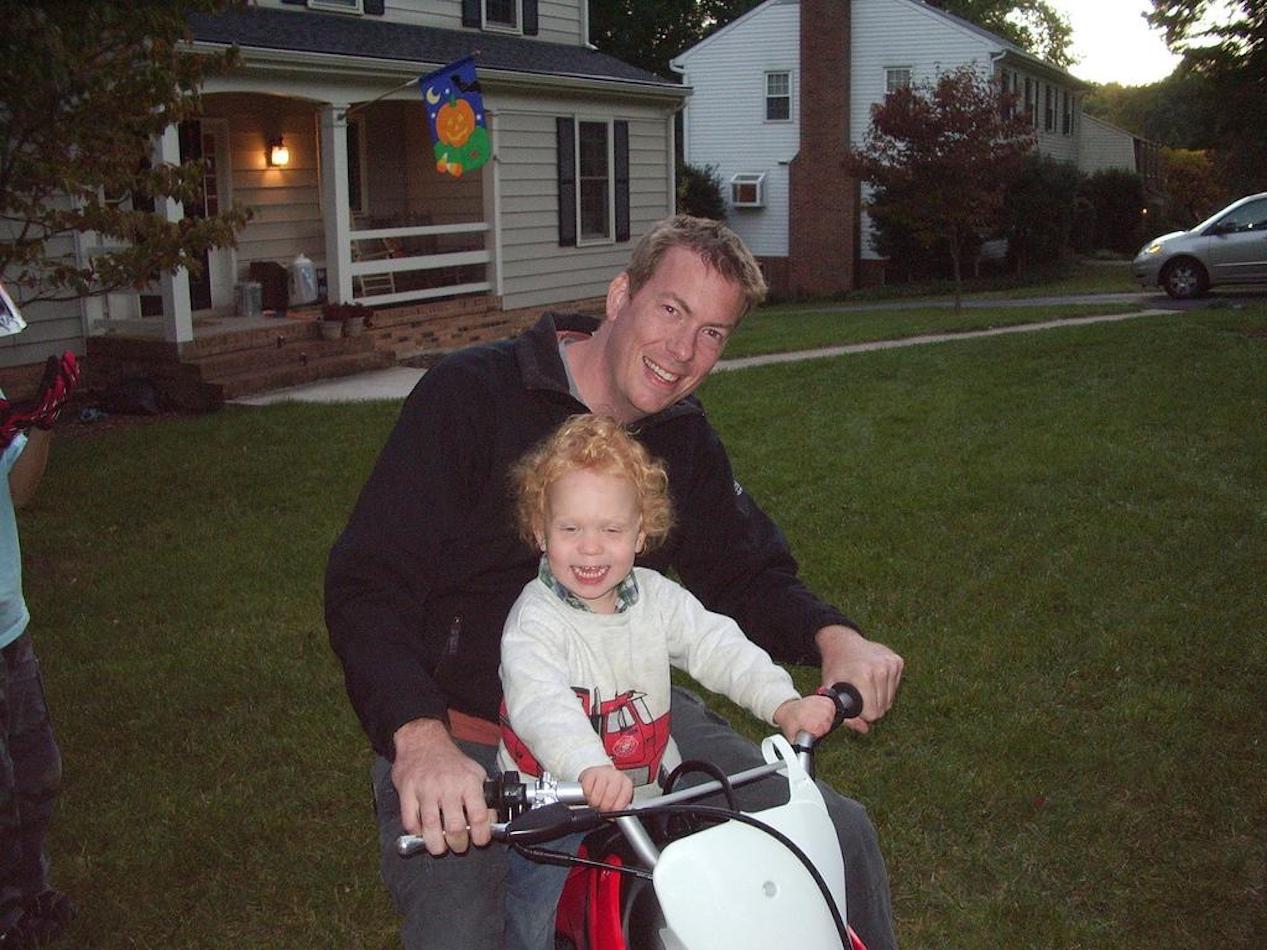}
    \vspace{-0.8cm}
    \caption*{Min: 3.381}
  \end{minipage}
  \hfill
  \begin{minipage}[h]{0.44\columnwidth}
    \includegraphics[width=\columnwidth]{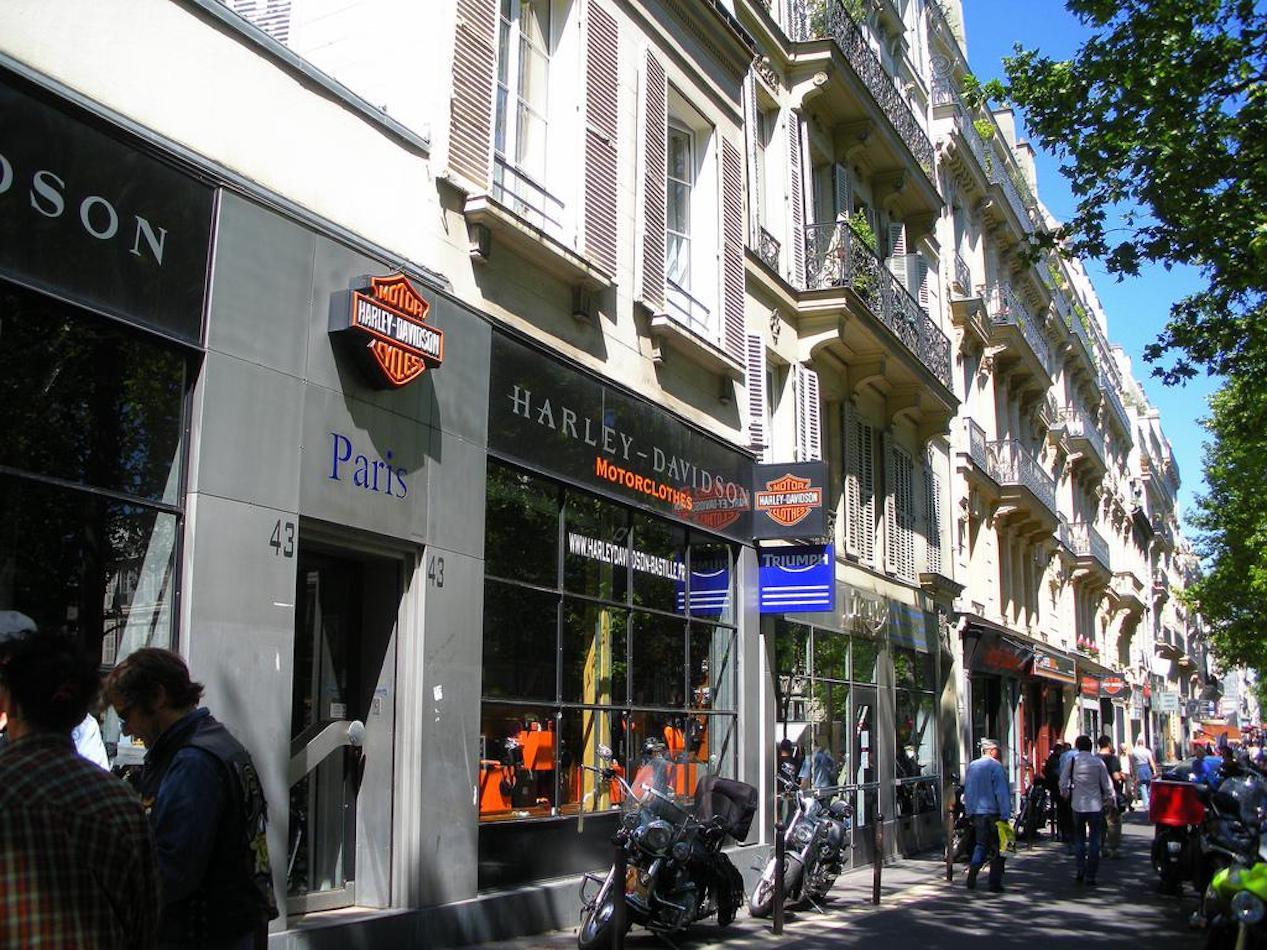}
    \vspace{-0.8cm}
    \caption*{Max: 3.488}
  \end{minipage}
    \caption{Images with the minimum and maximum predicted mean onsets. The image with the minimum was also predicted to elicit the lowest variation in gaze.}
    \label{fig:exp_pred_onset}
\end{figure}

\subsection{Examples}
\begin{figure}[h]
    \centering
    \begin{minipage}[h]{0.44\columnwidth}
    \includegraphics[width=\columnwidth]{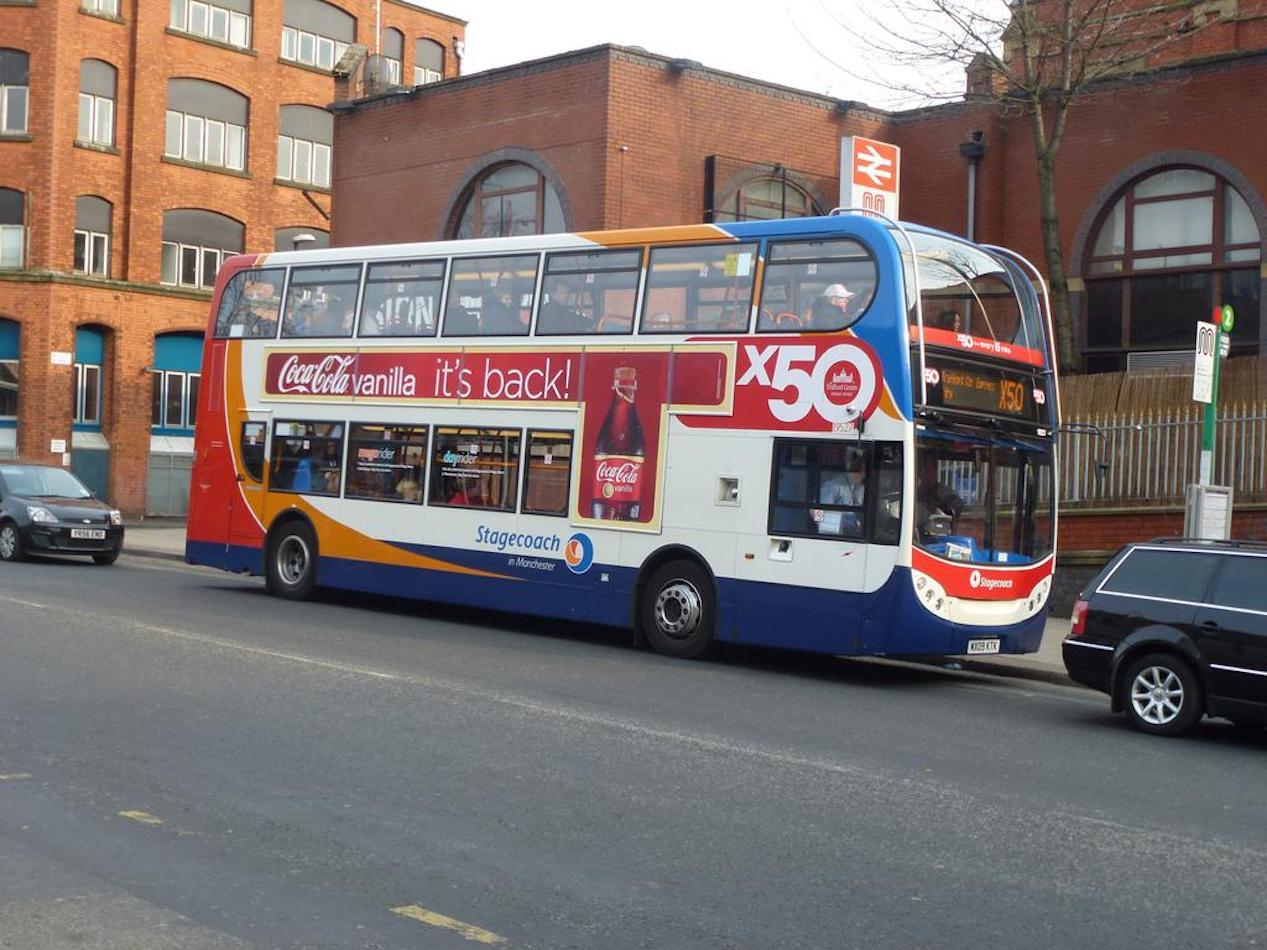}
    \vspace{-0.8cm}
    \caption*{Min: 0.529}
  \end{minipage}
  \hfill
  \begin{minipage}[h]{0.44\columnwidth}
    \includegraphics[width=\columnwidth]{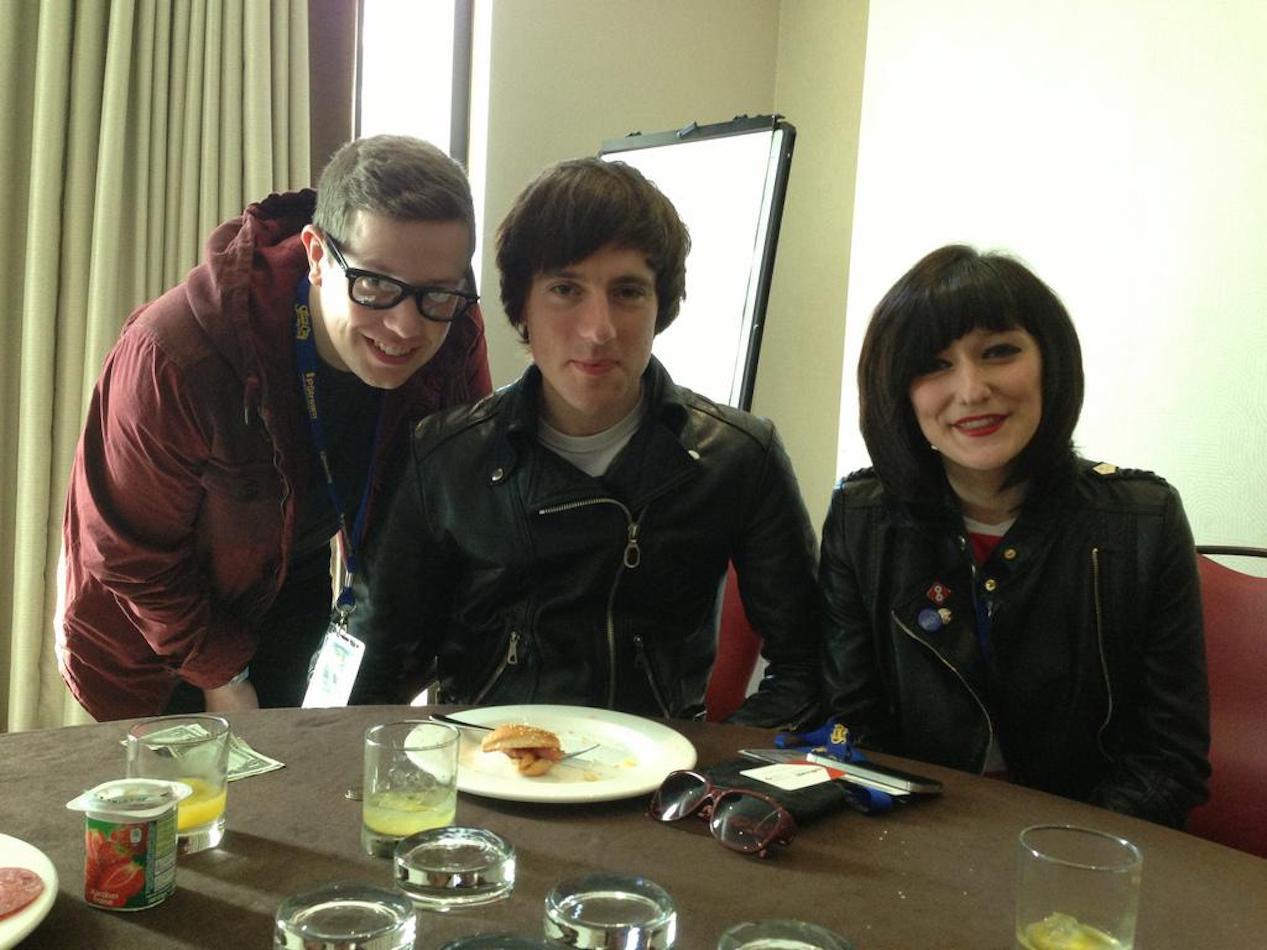}
    \vspace{-0.8cm}
    \caption*{Max: 0.541}
  \end{minipage}
    \caption{BLEU-2-based linguistic variation scores as predicted by the similarity-based approach.}
    \label{fig:exp_pred_imgspec}
\end{figure}

We illustrate the images with the minimum and maximum mean onsets as predicted by the similarity-based approach in Figure~\ref{fig:exp_pred_onset}. Figure~\ref{fig:exp_pred_imgspec} depicts predicted variation in descriptions, and Figure~\ref{fig:exp_pred_gazevar} the predicted variation in gaze.  %
We see a tendency to predict shorter speech onsets, more similar descriptions and gaze patterns in images containing a couple of people compared to scenes of streets with no visible or salient humans, a finding resonating with the conclusions drawn by~\citet{Jas_2015_CVPR}. This is potentially due to the salient and non-ambiguous nature of humans in images, as opposed to general street scenes with cars, buses and non-salient humans.

\begin{figure}[h]
    \centering
    \begin{minipage}[h]{0.44\columnwidth}
    \includegraphics[width=\columnwidth]{images/150458.jpg}
    \vspace{-0.8cm}
    \caption*{Min: 23.666}
  \end{minipage}
  \hfill
  \begin{minipage}[h]{0.44\columnwidth}
    \includegraphics[width=\columnwidth]{images/713742.jpg}
    \vspace{-0.8cm}
    \caption*{Max: 24.308}
  \end{minipage}
    \caption{Variation in gaze as predicted by the similarity-based approach.}
    \label{fig:exp_pred_gazevar}
\end{figure}

\section{Conclusion}
We quantified the variation in speech onsets, starting points, descriptions and gaze using a Dutch dataset of image descriptions with eye-tracking data. Our findings revealed the extent of variation in the process of describing images, and that variations in different signals correlate with each other. Furthermore, using a similarity-based prediction approach, we showed that image representations encoded by pretrained vision encoders capture variation in visuo-linguistic behavior to a weak-to-moderate extent. This pattern can be interpreted in light of models' pretraining objectives, as the predictions correlated more strongly for signals more internal to the objectives. Our study has implications for how human processes unfold as well as pretrained models' capabilities to represent such processes. %

Human and machine processing have differences, and we are motivated by the potential benefits of making the models increasingly knowledgeable about the multimodal landscape of human data. Although the impact of fine-tuning an already powerful pretrained model on a small-scale dataset with human signals could be modest, we hope that our work motivates the collection of more signals during crowdsourcing. For instance, it would be beneficial to take into account how long it took participants to complete a task given a certain stimulus, indicating the relative complexity and the uncertainty induced by the task as well as the stimulus. By inducing biases based on human signals, models can further take advantage of the information contained within such signals. Although it would be difficult to capture the full extent of the intricacies of human processing, this could help, for instance, a model interacting with human users to generate responses more aligned with human expectations.

\section*{Limitations}
In this work, we use a dataset in Dutch; however, the crossmodal interaction between vision and language could show some variation based on the properties of the languages (i.e. word order and morphological constraints), leading to variation in visual attention and structural choices~\citep{Norcliffe2015, myachykov}. Therefore, the findings might differ based on the languages of the datasets and the pretrained models. It would also be informative to explore other models and tasks, as well as explicit, discrete features that would contribute to the prediction of visuo-linguistic variation. Regarding the data, there could be possible noise in human signals and our preprocessing steps that affect the findings. Investigating the variation in gaze before/after speech onset with participant-specific analyses could also reveal interesting dynamics. As the dataset contains descriptions from 45 participants, with on average 15 participants describing each image, a different pool of participants (in particular, of a different size) may produce disparate results. A larger corpus may also allow for the training and fine-tuning of models. This is a line of work we have not explored in detail in this work, as a probing approach where we trained lightweight layers on top of image representations yielded even lower correlation coefficients and higher losses.

\section*{Ethics Statement}
The data we use in this work had been collected following ethical guidelines~\citep{miltenburg2018DIDEC}. Predicting or using eye-tracking of humans in the wild would require ethical approval, a line of research which we do not investigate, as our focus is on whether pretrained model representations can account for variations in human outputs. Since we use large pretrained models in frozen form, they may be perpetuating biases that are not desirable.

\section*{Acknowledgements}
We would like to thank the members of the Dialogue Modelling Group for their valuable feedback regarding the project, and Andrea Bruera for his pointers to the work by \citet{ANDERSON201644}. We also would like to thank the anonymous ARR reviewers for their thoughtful and useful feedback. This research has received funding from the European Research Council (ERC) under the European Union’s Horizon 2020 research and innovation programme (grant agreement No. 819455).

\bibliography{custom}

\begin{thebibliography}{85}
\expandafter\ifx\csname natexlab\endcsname\relax\def\natexlab#1{#1}\fi

\bibitem[{Alacam et~al.(2022)Alacam, Ruppert, Malhotra, Biemann, and Zarrie{\ss}}]{alacam-etal-2022-modeling-updated}
{\"O}zge Alacam, Eugen Ruppert, Ganeshan Malhotra, Chris Biemann, and Sina Zarrie{\ss}. 2022.
\newblock \href {https://aclanthology.org/2022.findings-aacl.19} {Modeling referential gaze in task-oriented settings of varying referential complexity}.
\newblock In \emph{Findings of the Association for Computational Linguistics: AACL-IJCNLP 2022}, pages 197--210, Online only. Association for Computational Linguistics.

\bibitem[{Alayrac et~al.(2022)Alayrac, Donahue, Luc, Miech, Barr, Hasson, Lenc, Mensch, Millican, Reynolds, Ring, Rutherford, Cabi, Han, Gong, Samangooei, Monteiro, Menick, Borgeaud, Brock, Nematzadeh, Sharifzadeh, Bi\'{n}kowski, Barreira, Vinyals, Zisserman, and Simonyan}]{flamingo}
Jean-Baptiste Alayrac, Jeff Donahue, Pauline Luc, Antoine Miech, Iain Barr, Yana Hasson, Karel Lenc, Arthur Mensch, Katherine Millican, Malcolm Reynolds, Roman Ring, Eliza Rutherford, Serkan Cabi, Tengda Han, Zhitao Gong, Sina Samangooei, Marianne Monteiro, Jacob~L Menick, Sebastian Borgeaud, Andy Brock, Aida Nematzadeh, Sahand Sharifzadeh, Miko{\l}aj Bi\'{n}kowski, Ricardo Barreira, Oriol Vinyals, Andrew Zisserman, and Kar\'{e}n Simonyan. 2022.
\newblock \href {https://proceedings.neurips.cc/paper_files/paper/2022/file/960a172bc7fbf0177ccccbb411a7d800-Paper-Conference.pdf} {Flamingo: a visual language model for few-shot learning}.
\newblock In \emph{Advances in Neural Information Processing Systems}, volume~35, pages 23716--23736. Curran Associates, Inc.

\bibitem[{Anderson et~al.(2017)Anderson, Kiela, Clark, and Poesio}]{anderson-etal-2017-visually}
Andrew~J. Anderson, Douwe Kiela, Stephen Clark, and Massimo Poesio. 2017.
\newblock \href {https://doi.org/10.1162/tacl_a_00043} {Visually grounded and textual semantic models differentially decode brain activity associated with concrete and abstract nouns}.
\newblock \emph{Transactions of the Association for Computational Linguistics}, 5:17--30.

\bibitem[{Anderson et~al.(2016)Anderson, Zinszer, and Raizada}]{ANDERSON201644}
Andrew~James Anderson, Benjamin~D. Zinszer, and Rajeev~D.S. Raizada. 2016.
\newblock \href {https://doi.org/https://doi.org/10.1016/j.neuroimage.2015.12.035} {Representational similarity encoding for fmri: Pattern-based synthesis to predict brain activity using stimulus-model-similarities}.
\newblock \emph{NeuroImage}, 128:44--53.

\bibitem[{Arehalli et~al.(2022)Arehalli, Dillon, and Linzen}]{arehalli-etal-2022-syntactic}
Suhas Arehalli, Brian Dillon, and Tal Linzen. 2022.
\newblock \href {https://aclanthology.org/2022.conll-1.20} {Syntactic surprisal from neural models predicts, but underestimates, human processing difficulty from syntactic ambiguities}.
\newblock In \emph{Proceedings of the 26th Conference on Computational Natural Language Learning (CoNLL)}, pages 301--313, Abu Dhabi, United Arab Emirates (Hybrid). Association for Computational Linguistics.

\bibitem[{Bain et~al.(2023)Bain, Huh, Han, and Zisserman}]{bain2022whisperx}
Max Bain, Jaesung Huh, Tengda Han, and Andrew Zisserman. 2023.
\newblock Whisperx: Time-accurate speech transcription of long-form audio.
\newblock \emph{INTERSPEECH 2023}.

\bibitem[{Barrett et~al.(2018)Barrett, Bingel, Hollenstein, Rei, and S{\o}gaard}]{barrett-etal-2018-sequence}
Maria Barrett, Joachim Bingel, Nora Hollenstein, Marek Rei, and Anders S{\o}gaard. 2018.
\newblock \href {https://doi.org/10.18653/v1/K18-1030} {Sequence classification with human attention}.
\newblock In \emph{Proceedings of the 22nd Conference on Computational Natural Language Learning}, pages 302--312, Brussels, Belgium. Association for Computational Linguistics.

\bibitem[{Barrett et~al.(2016)Barrett, Bingel, Keller, and S{\o}gaard}]{barrett-etal-2016-weakly}
Maria Barrett, Joachim Bingel, Frank Keller, and Anders S{\o}gaard. 2016.
\newblock \href {https://doi.org/10.18653/v1/P16-2094} {Weakly supervised part-of-speech tagging using eye-tracking data}.
\newblock In \emph{Proceedings of the 54th Annual Meeting of the Association for Computational Linguistics (Volume 2: Short Papers)}, pages 579--584, Berlin, Germany. Association for Computational Linguistics.

\bibitem[{Berger et~al.(2023)Berger, Frermann, Stanovsky, and Abend}]{constraints}
Uri Berger, Lea Frermann, Gabriel Stanovsky, and Omri Abend. 2023.
\newblock \href {https://aclanthology.org/2023.findings-eacl.172} {A large-scale multilingual study of visual constraints on linguistic selection of descriptions}.
\newblock In \emph{Findings of the Association for Computational Linguistics: EACL 2023}, pages 2285--2299, Dubrovnik, Croatia. Association for Computational Linguistics.

\bibitem[{Berrios et~al.(2023)Berrios, Mittal, Thrush, Kiela, and Singh}]{lens}
William Berrios, Gautam Mittal, Tristan Thrush, Douwe Kiela, and Amanpreet Singh. 2023.
\newblock \href {http://arxiv.org/abs/2306.16410} {Towards language models that can see: Computer vision through the lens of natural language}.

\bibitem[{Bock et~al.(2004)Bock, Irwin, and Davidson}]{putting}
Kathryn Bock, David~E. Irwin, and Douglas~J. Davidson. 2004.
\newblock Putting first things first.
\newblock \emph{The interface of language, vision, and action: Eye movements and the visual world}, pages 249--278.

\bibitem[{Bruera and Poesio(2023)}]{Bruera2023.08.23.554436}
Andrea Bruera and Massimo Poesio. 2023.
\newblock \href {https://doi.org/10.1101/2023.08.23.554436} {Family lexicon: using language models to encode memories of personally familiar and famous people and places in the brain}.
\newblock \emph{bioRxiv}.

\bibitem[{Bruera et~al.(2023)Bruera, Tao, Anderson, Çokal, Haber, and Poesio}]{https://doi.org/10.1111/cogs.13388}
Andrea Bruera, Yuan Tao, Andrew Anderson, Derya Çokal, Janosch Haber, and Massimo Poesio. 2023.
\newblock \href {https://doi.org/https://doi.org/10.1111/cogs.13388} {Modeling brain representations of words' concreteness in context using gpt-2 and human ratings}.
\newblock \emph{Cognitive Science}, 47(12):e13388.

\bibitem[{Buswell(1935)}]{buswell1935people}
Guy~Thomas Buswell. 1935.
\newblock \emph{How people look at pictures: A study of the psychology and perception in art}.
\newblock University of Chicago Press.

\bibitem[{Castelhano et~al.(2009)Castelhano, Mack, and Henderson}]{castelhano}
Monica~S. Castelhano, Michael~L. Mack, and John~M. Henderson. 2009.
\newblock \href {https://doi.org/10.1167/9.3.6} {{Viewing task influences eye movement control during active scene perception}}.
\newblock \emph{Journal of Vision}, 9(3):6--6.

\bibitem[{Chen et~al.(2022)Chen, Guo, Yi, Li, and Elhoseiny}]{visualgpt}
Jun Chen, Han Guo, Kai Yi, Boyang Li, and Mohamed Elhoseiny. 2022.
\newblock Visualgpt: Data-efficient adaptation of pretrained language models for image captioning.
\newblock In \emph{Proceedings of the IEEE/CVF Conference on Computer Vision and Pattern Recognition (CVPR)}, pages 18030--18040.

\bibitem[{Coco and Keller(2012)}]{scanpatterns}
Moreno~I. Coco and Frank Keller. 2012.
\newblock \href {https://doi.org/10.1111/j.1551-6709.2012.01246.x} {Scan patterns predict sentence production in the cross-modal processing of visual scenes}.
\newblock \emph{Cognitive Science}, 36(7):1204--1223.

\bibitem[{Coco and Keller(2015)}]{integrating}
Moreno~I. Coco and Frank Keller. 2015.
\newblock \href {https://doi.org/10.1007/s10339-014-0642-0} {Integrating mechanisms of visual guidance in naturalistic language production}.
\newblock \emph{Cognitive processing}, 16(2):131—150.

\bibitem[{Conneau et~al.(2018)Conneau, Kruszewski, Lample, Barrault, and Baroni}]{conneau-etal-2018-cram}
Alexis Conneau, German Kruszewski, Guillaume Lample, Lo{\"\i}c Barrault, and Marco Baroni. 2018.
\newblock \href {https://doi.org/10.18653/v1/P18-1198} {What you can cram into a single {\$}{\&}!{\#}* vector: Probing sentence embeddings for linguistic properties}.
\newblock In \emph{Proceedings of the 56th Annual Meeting of the Association for Computational Linguistics (Volume 1: Long Papers)}, pages 2126--2136, Melbourne, Australia. Association for Computational Linguistics.

\bibitem[{Das et~al.(2016)Das, Agrawal, Zitnick, Parikh, and Batra}]{das-etal-2016-human}
Abhishek Das, Harsh Agrawal, Larry Zitnick, Devi Parikh, and Dhruv Batra. 2016.
\newblock \href {https://doi.org/10.18653/v1/D16-1092} {Human attention in visual question answering: Do humans and deep networks look at the same regions?}
\newblock In \emph{Proceedings of the 2016 Conference on Empirical Methods in Natural Language Processing}, pages 932--937, Austin, Texas. Association for Computational Linguistics.

\bibitem[{de~Vries et~al.(2019)de~Vries, van Cranenburgh, Bisazza, Caselli, Noord, and Nissim}]{devries2019bertje}
Wietse de~Vries, Andreas van Cranenburgh, Arianna Bisazza, Tommaso Caselli, Gertjan~van Noord, and Malvina Nissim. 2019.
\newblock \href {http://arxiv.org/abs/1912.09582} {{BERTje}: {A} {Dutch} {BERT} {Model}}.
\newblock arXiv:1912.09582.

\bibitem[{Ding et~al.(2022)Ding, Chen, Du, Qin, and Liu}]{cogbert}
Xiao Ding, Bowen Chen, Li~Du, Bing Qin, and Ting Liu. 2022.
\newblock \href {https://aclanthology.org/2022.coling-1.284} {{C}og{BERT}: Cognition-guided pre-trained language models}.
\newblock In \emph{Proceedings of the 29th International Conference on Computational Linguistics}, pages 3210--3225, Gyeongju, Republic of Korea. International Committee on Computational Linguistics.

\bibitem[{Dong et~al.(2022)Dong, Goldstein, and Yang}]{gazby}
Sibo Dong, Justin Goldstein, and Grace~Hui Yang. 2022.
\newblock \href {https://doi.org/10.1145/3539813.3545129} {Gazby: Gaze-based bert model to incorporate human attention in neural information retrieval}.
\newblock In \emph{Proceedings of the 2022 ACM SIGIR International Conference on Theory of Information Retrieval}, ICTIR '22, page 182–192, New York, NY, USA. Association for Computing Machinery.

\bibitem[{Dosovitskiy et~al.(2021)Dosovitskiy, Beyer, Kolesnikov, Weissenborn, Zhai, Unterthiner, Dehghani, Minderer, Heigold, Gelly, Uszkoreit, and Houlsby}]{dosovitskiy2021an}
Alexey Dosovitskiy, Lucas Beyer, Alexander Kolesnikov, Dirk Weissenborn, Xiaohua Zhai, Thomas Unterthiner, Mostafa Dehghani, Matthias Minderer, Georg Heigold, Sylvain Gelly, Jakob Uszkoreit, and Neil Houlsby. 2021.
\newblock \href {https://openreview.net/forum?id=YicbFdNTTy} {An image is worth 16x16 words: Transformers for image recognition at scale}.
\newblock In \emph{International Conference on Learning Representations}.

\bibitem[{Esaulova et~al.(2019)Esaulova, Penke, and Dolscheid}]{esaulova2019DescribingEC}
Yulia Esaulova, Martina Penke, and Sarah Dolscheid. 2019.
\newblock \href {https://api.semanticscholar.org/CorpusID:116875999} {Describing events: Changes in eye movements and language production due to visual and conceptual properties of scenes}.
\newblock \emph{Frontiers in Psychology}, 10.

\bibitem[{Ferreira and Rehrig(2019)}]{linear}
Fernanda Ferreira and Gwendolyn Rehrig. 2019.
\newblock Linearisation during language production: evidence from scene meaning and saliency maps.
\newblock \emph{Language, Cognition and Neuroscience}, 34(9):1129--1139.

\bibitem[{Gatt et~al.(2017)Gatt, Krahmer, van Deemter, and van Gompel}]{gatt}
Albert Gatt, Emiel Krahmer, Kees van Deemter, and Roger~P.G. van Gompel. 2017.
\newblock \href {https://doi.org/https://doi.org/10.1111/cogs.12375} {Reference production as search: The impact of domain size on the production of distinguishing descriptions}.
\newblock \emph{Cognitive Science}, 41(S6):1457--1492.

\bibitem[{Gella and Keller(2018)}]{gella-keller-2018-evaluation}
Spandana Gella and Frank Keller. 2018.
\newblock \href {https://doi.org/10.18653/v1/N18-2119} {An evaluation of image-based verb prediction models against human eye-tracking data}.
\newblock In \emph{Proceedings of the 2018 Conference of the North American Chapter of the Association for Computational Linguistics: Human Language Technologies, Volume 2 (Short Papers)}, pages 758--763, New Orleans, Louisiana. Association for Computational Linguistics.

\bibitem[{Gleitman et~al.(2007)Gleitman, January, Nappa, and Trueswell}]{gleitman2007give}
Lila~R. Gleitman, David January, Rebecca Nappa, and John~C. Trueswell. 2007.
\newblock On the give and take between event apprehension and utterance formulation.
\newblock \emph{Journal of Memory and Language}, 57(4):544--569.

\bibitem[{Griffin(2004)}]{whylook}
Zenzi~M. Griffin. 2004.
\newblock Why look? reasons for eye movements related to language production.
\newblock In J.~M. Henderson \&~F. Ferreira, editor, \emph{The interface of language, vision, and action: Eye movements and the visual world}, chapter~7, pages 213--247. Psychology Press, New York.

\bibitem[{Griffin and Bock(2000)}]{griffinbock}
Zenzi~M. Griffin and Kathryn Bock. 2000.
\newblock \href {https://doi.org/10.1111/1467-9280.00255} {What the eyes say about speaking}.
\newblock \emph{Psychological Science}, 11:274--9.

\bibitem[{Gualdoni et~al.(2023)Gualdoni, Brochhagen, Mädebach, and Boleda}]{GUALDONI2023104459}
Eleonora Gualdoni, Thomas Brochhagen, Andreas Mädebach, and Gemma Boleda. 2023.
\newblock \href {https://doi.org/https://doi.org/10.1016/j.jml.2023.104459} {What’s in a name? a large-scale computational study on how competition between names affects naming variation}.
\newblock \emph{Journal of Memory and Language}, 133:104459.

\bibitem[{He et~al.(2019)He, Tavakoli, Borji, and Pugeault}]{He2019HumanAI}
Sen He, Hamed~R. Tavakoli, Ali Borji, and Nicolas Pugeault. 2019.
\newblock Human attention in image captioning: Dataset and analysis.
\newblock \emph{2019 IEEE/CVF International Conference on Computer Vision (ICCV)}, pages 8528--8537.

\bibitem[{Henderson(2017)}]{HENDERSON201715}
John~M. Henderson. 2017.
\newblock \href {https://doi.org/https://doi.org/10.1016/j.tics.2016.11.003} {Gaze control as prediction}.
\newblock \emph{Trends in Cognitive Sciences}, 21(1):15--23.

\bibitem[{Henderson and Ferreira(2013)}]{henderson2013interface}
John~M. Henderson and Fernanda Ferreira. 2013.
\newblock \emph{The Interface of Language, Vision, and Action: Eye Movements and the Visual World}.
\newblock Taylor \& Francis.

\bibitem[{Hollenstein et~al.(2022)Hollenstein, Chersoni, Jacobs, Oseki, Pr{\'e}vot, and Santus}]{hollenstein-etal-2022-cmcl}
Nora Hollenstein, Emmanuele Chersoni, Cassandra Jacobs, Yohei Oseki, Laurent Pr{\'e}vot, and Enrico Santus. 2022.
\newblock \href {https://doi.org/10.18653/v1/2022.cmcl-1.14} {{CMCL} 2022 shared task on multilingual and crosslingual prediction of human reading behavior}.
\newblock In \emph{Proceedings of the Workshop on Cognitive Modeling and Computational Linguistics}, pages 121--129, Dublin, Ireland. Association for Computational Linguistics.

\bibitem[{Hollenstein et~al.(2021{\natexlab{a}})Hollenstein, Chersoni, Jacobs, Oseki, Pr{\'e}vot, and Santus}]{hollenstein-etal-2021-cmcl}
Nora Hollenstein, Emmanuele Chersoni, Cassandra~L. Jacobs, Yohei Oseki, Laurent Pr{\'e}vot, and Enrico Santus. 2021{\natexlab{a}}.
\newblock \href {https://doi.org/10.18653/v1/2021.cmcl-1.7} {{CMCL} 2021 shared task on eye-tracking prediction}.
\newblock In \emph{Proceedings of the Workshop on Cognitive Modeling and Computational Linguistics}, pages 72--78, Online. Association for Computational Linguistics.

\bibitem[{Hollenstein et~al.(2021{\natexlab{b}})Hollenstein, Pirovano, Zhang, J{\"a}ger, and Beinborn}]{hollenstein-etal-2021-multilingual}
Nora Hollenstein, Federico Pirovano, Ce~Zhang, Lena J{\"a}ger, and Lisa Beinborn. 2021{\natexlab{b}}.
\newblock \href {https://doi.org/10.18653/v1/2021.naacl-main.10} {Multilingual language models predict human reading behavior}.
\newblock In \emph{Proceedings of the 2021 Conference of the North American Chapter of the Association for Computational Linguistics: Human Language Technologies}, pages 106--123, Online. Association for Computational Linguistics.

\bibitem[{Jas and Parikh(2015)}]{Jas_2015_CVPR}
Mainak Jas and Devi Parikh. 2015.
\newblock \href {https://api.semanticscholar.org/CorpusID:14088994} {Image specificity}.
\newblock In \emph{2015 IEEE Conference on Computer Vision and Pattern Recognition (CVPR)}, pages 2727--2736.

\bibitem[{Kahneman(2012)}]{kahneman}
Daniel Kahneman. 2012.
\newblock \emph{Thinking, Fast and Slow}.
\newblock Penguin Books.

\bibitem[{Khurana et~al.(2023)Khurana, Kumar, Hollenstein, Kumar, and Krishnamurthy}]{synthesizing}
Varun Khurana, Yaman Kumar, Nora Hollenstein, Rajesh Kumar, and Balaji Krishnamurthy. 2023.
\newblock \href {https://aclanthology.org/2023.eacl-main.139} {Synthesizing human gaze feedback for improved {NLP} performance}.
\newblock In \emph{Proceedings of the 17th Conference of the European Chapter of the Association for Computational Linguistics}, pages 1895--1908, Dubrovnik, Croatia. Association for Computational Linguistics.

\bibitem[{Kirillov et~al.(2023)Kirillov, Mintun, Ravi, Mao, Rolland, Gustafson, Xiao, Whitehead, Berg, Lo, Dollar, and Girshick}]{segm}
Alexander Kirillov, Eric Mintun, Nikhila Ravi, Hanzi Mao, Chloe Rolland, Laura Gustafson, Tete Xiao, Spencer Whitehead, Alexander~C. Berg, Wan-Yen Lo, Piotr Dollar, and Ross Girshick. 2023.
\newblock Segment anything.
\newblock In \emph{Proceedings of the IEEE/CVF International Conference on Computer Vision (ICCV)}, pages 4015--4026.

\bibitem[{Klerke et~al.(2016)Klerke, Goldberg, and S{\o}gaard}]{klerke-etal-2016-improving}
Sigrid Klerke, Yoav Goldberg, and Anders S{\o}gaard. 2016.
\newblock \href {https://doi.org/10.18653/v1/N16-1179} {Improving sentence compression by learning to predict gaze}.
\newblock In \emph{Proceedings of the 2016 Conference of the North {A}merican Chapter of the Association for Computational Linguistics: Human Language Technologies}, pages 1528--1533, San Diego, California. Association for Computational Linguistics.

\bibitem[{Levelt(1981)}]{levelt}
Willem J.~M. Levelt. 1981.
\newblock \href {https://api.semanticscholar.org/CorpusID:19942842} {The speaker's linearization problem}.
\newblock \emph{Philosophical Transactions of the Royal Society B}, 295:305--315.

\bibitem[{Li et~al.(2023)Li, Li, Savarese, and Hoi}]{li2023blip2}
Junnan Li, Dongxu Li, Silvio Savarese, and Steven Hoi. 2023.
\newblock \href {https://proceedings.mlr.press/v202/li23q.html} {{BLIP}-2: Bootstrapping language-image pre-training with frozen image encoders and large language models}.
\newblock In \emph{Proceedings of the 40th International Conference on Machine Learning}, volume 202 of \emph{Proceedings of Machine Learning Research}, pages 19730--19742. PMLR.

\bibitem[{Lin et~al.(2014)Lin, Maire, Belongie, Hays, Perona, Ramanan, Doll{\'a}r, and Zitnick}]{coco}
Tsung-Yi Lin, Michael Maire, Serge Belongie, James Hays, Pietro Perona, Deva Ramanan, Piotr Doll{\'a}r, and C.~Lawrence Zitnick. 2014.
\newblock Microsoft {COCO}: Common objects in context.
\newblock In \emph{Computer Vision -- ECCV 2014}, pages 740--755, Cham. Springer International Publishing.

\bibitem[{MacWhinney(1977)}]{starting}
Brian MacWhinney. 1977.
\newblock \href {http://www.jstor.org/stable/413059} {Starting points}.
\newblock \emph{Language}, 53(1):152--168.

\bibitem[{Ma{\~n}as et~al.(2023)Ma{\~n}as, Rodriguez~Lopez, Ahmadi, Nematzadeh, Goyal, and Agrawal}]{mapl}
Oscar Ma{\~n}as, Pau Rodriguez~Lopez, Saba Ahmadi, Aida Nematzadeh, Yash Goyal, and Aishwarya Agrawal. 2023.
\newblock \href {https://aclanthology.org/2023.eacl-main.185} {{MAPL}: Parameter-efficient adaptation of unimodal pre-trained models for vision-language few-shot prompting}.
\newblock In \emph{Proceedings of the 17th Conference of the European Chapter of the Association for Computational Linguistics}, pages 2523--2548, Dubrovnik, Croatia. Association for Computational Linguistics.

\bibitem[{Mathias et~al.(2020)Mathias, Kanojia, Mishra, and Bhattacharya}]{ijcai2020p0683}
Sandeep Mathias, Diptesh Kanojia, Abhijit Mishra, and Pushpak Bhattacharya. 2020.
\newblock \href {https://doi.org/10.24963/ijcai.2020/683} {A survey on using gaze behaviour for natural language processing}.
\newblock In \emph{Proceedings of the Twenty-Ninth International Joint Conference on Artificial Intelligence, {IJCAI-20}}, pages 4907--4913. International Joint Conferences on Artificial Intelligence Organization.
\newblock Survey track.

\bibitem[{Meyer(2004)}]{meyer}
Antje~S. Meyer. 2004.
\newblock The use of eye tracking in studies of sentence generation.
\newblock In J.~M. Henderson and F.~Ferreira, editors, \emph{The interface of language, vision, and action: Eye movements and the visual world}, chapter~6, pages 191--212. Psychology Press, New York.

\bibitem[{Meyer and van~der Meulen(2000)}]{Meyer2000PhonologicalPE}
Antje~S. Meyer and Femke van~der Meulen. 2000.
\newblock Phonological priming effects on speech onset latencies and viewing times in object naming.
\newblock \emph{Psychonomic Bulletin \& Review}, 7:314--319.

\bibitem[{Mikolov et~al.(2013)Mikolov, Chen, Corrado, and Dean}]{mikolov2013efficient}
Tomas Mikolov, Kai Chen, Greg Corrado, and Jeffrey Dean. 2013.
\newblock \href {http://arxiv.org/abs/1301.3781} {Efficient estimation of word representations in vector space}.

\bibitem[{Miller(1994)}]{miller-1994-wordnet}
George~A. Miller. 1994.
\newblock \href {https://aclanthology.org/H94-1111} {{W}ord{N}et: A lexical database for {E}nglish}.
\newblock In \emph{{H}uman {L}anguage {T}echnology: Proceedings of a Workshop held at {P}lainsboro, {N}ew {J}ersey, {M}arch 8-11, 1994}.

\bibitem[{Mishra and Bhattacharyya(2018)}]{mishra}
Abhijit Mishra and Pushpak Bhattacharyya. 2018.
\newblock \emph{Cognitively Inspired Natural Language Processing: An Investigation Based on Eye-Tracking}, 1st edition.
\newblock Springer Publishing Company, Incorporated.

\bibitem[{Mokady et~al.(2021)Mokady, Hertz, and Bermano}]{mokady2021clipcap}
Ron Mokady, Amir Hertz, and Amit~H. Bermano. 2021.
\newblock Clipcap: Clip prefix for image captioning.
\newblock \emph{arXiv preprint arXiv:2111.09734}.

\bibitem[{Myachykov et~al.(2011)Myachykov, Thompson, Scheepers, and Garrod}]{myachykov}
Andriy Myachykov, Dominic Thompson, Christoph Scheepers, and Simon Garrod. 2011.
\newblock \href {https://doi.org/https://doi.org/10.1111/j.1749-818X.2010.00265.x} {Visual attention and structural choice in sentence production across languages}.
\newblock \emph{Language and Linguistics Compass}, 5(2):95--107.

\bibitem[{Norcliffe and Konopka(2015)}]{Norcliffe2015}
Elisabeth Norcliffe and Agnieszka~E. Konopka. 2015.
\newblock \href {https://doi.org/10.1007/978-81-322-2443-3_5} {\emph{Vision and Language in Cross-Linguistic Research on Sentence Production}}, pages 77--96. Springer India, New Delhi.

\bibitem[{Oh and Schuler(2023{\natexlab{a}})}]{oh2023transformerbased}
Byung-Doh Oh and William Schuler. 2023{\natexlab{a}}.
\newblock \href {https://doi.org/10.18653/v1/2023.findings-emnlp.128} {Transformer-based language model surprisal predicts human reading times best with about two billion training tokens}.
\newblock In \emph{Findings of the Association for Computational Linguistics: EMNLP 2023}, pages 1915--1921, Singapore. Association for Computational Linguistics.

\bibitem[{Oh and Schuler(2023{\natexlab{b}})}]{10.1162/tacl_a_00548}
Byung-Doh Oh and William Schuler. 2023{\natexlab{b}}.
\newblock \href {https://doi.org/10.1162/tacl_a_00548} {{Why Does Surprisal From Larger Transformer-Based Language Models Provide a Poorer Fit to Human Reading Times?}}
\newblock \emph{Transactions of the Association for Computational Linguistics}, 11:336--350.

\bibitem[{Oliva(2005)}]{oliva2005gist}
Aude Oliva. 2005.
\newblock Gist of the scene.
\newblock In \emph{Neurobiology of attention}, pages 251--256. Elsevier.

\bibitem[{Oliva and Torralba(2006)}]{oliva2006building}
Aude Oliva and Antonio Torralba. 2006.
\newblock Building the gist of a scene: The role of global image features in recognition.
\newblock \emph{Progress in brain research}, 155:23--36.

\bibitem[{Papineni et~al.(2002)Papineni, Roukos, Ward, and Zhu}]{Papineni:2002}
Kishore Papineni, Salim Roukos, Todd Ward, and Wei-Jing Zhu. 2002.
\newblock \href {https://doi.org/10.3115/1073083.1073135} {{BLEU}: A method for automatic evaluation of machine translation}.
\newblock In \emph{Proceedings of the 40th Annual Meeting on Association for Computational Linguistics}, pages 311--318. Association for Computational Linguistics.

\bibitem[{Pouw et~al.(2023)Pouw, Hollenstein, and Beinborn}]{pouw-etal-2023-cross}
Charlotte Pouw, Nora Hollenstein, and Lisa Beinborn. 2023.
\newblock \href {https://aclanthology.org/2023.findings-eacl.49} {Cross-lingual transfer of cognitive processing complexity}.
\newblock In \emph{Findings of the Association for Computational Linguistics: EACL 2023}, pages 655--669, Dubrovnik, Croatia. Association for Computational Linguistics.

\bibitem[{Prion and Haerling(2014)}]{Prion2014MakingSO}
Susan~K. Prion and Katie~Anne Haerling. 2014.
\newblock \href {https://api.semanticscholar.org/CorpusID:70494185} {Making sense of methods and measurement: Spearman-rho ranked-order correlation coefficient}.
\newblock \emph{Clinical Simulation in Nursing}, 10:535--536.

\bibitem[{Radford et~al.(2021)Radford, Kim, Hallacy, Ramesh, Goh, Agarwal, Sastry, Askell, Mishkin, Clark, Krueger, and Sutskever}]{clip}
Alec Radford, Jong~Wook Kim, Chris Hallacy, Aditya Ramesh, Gabriel Goh, Sandhini Agarwal, Girish Sastry, Amanda Askell, Pamela Mishkin, Jack Clark, Gretchen Krueger, and Ilya Sutskever. 2021.
\newblock \href {https://proceedings.mlr.press/v139/radford21a.html} {Learning transferable visual models from natural language supervision}.
\newblock In \emph{Proceedings of the 38th International Conference on Machine Learning}, volume 139 of \emph{Proceedings of Machine Learning Research}, pages 8748--8763. PMLR.

\bibitem[{Radford et~al.(2023)Radford, Kim, Xu, Brockman, McLeavey, and Sutskever}]{radford2022robust}
Alec Radford, Jong~Wook Kim, Tao Xu, Greg Brockman, Christine McLeavey, and Ilya Sutskever. 2023.
\newblock Robust speech recognition via large-scale weak supervision.
\newblock In \emph{Proceedings of the 40th International Conference on Machine Learning}, ICML'23. JMLR.org.

\bibitem[{Ren and Xiong(2021)}]{cogalign}
Yuqi Ren and Deyi Xiong. 2021.
\newblock \href {https://doi.org/10.18653/v1/2021.acl-long.291} {{C}og{A}lign: Learning to align textual neural representations to cognitive language processing signals}.
\newblock In \emph{Proceedings of the 59th Annual Meeting of the Association for Computational Linguistics and the 11th International Joint Conference on Natural Language Processing (Volume 1: Long Papers)}, pages 3758--3769, Online. Association for Computational Linguistics.

\bibitem[{Salvucci and Goldberg(2000)}]{salvucci}
Dario~D. Salvucci and Joseph~H. Goldberg. 2000.
\newblock \href {https://doi.org/10.1145/355017.355028} {Identifying fixations and saccades in eye-tracking protocols}.
\newblock In \emph{Proceedings of the 2000 Symposium on Eye Tracking Research \& Applications}, ETRA '00, page 71–78, New York, NY, USA. Association for Computing Machinery.

\bibitem[{Shen et~al.(2022)Shen, Li, Tan, Bansal, Rohrbach, Chang, Yao, and Keutzer}]{shen2022how}
Sheng Shen, Liunian~Harold Li, Hao Tan, Mohit Bansal, Anna Rohrbach, Kai-Wei Chang, Zhewei Yao, and Kurt Keutzer. 2022.
\newblock \href {https://openreview.net/forum?id=zf_Ll3HZWgy} {How much can {CLIP} benefit vision-and-language tasks?}
\newblock In \emph{International Conference on Learning Representations}.

\bibitem[{Slobin(2003)}]{Slobin2003-SLOLAT}
Dan~I. Slobin. 2003.
\newblock \href {https://doi.org/10.7551/mitpress/4117.003.0013} {{Language and Thought Online: Cognitive Consequences of Linguistic Relativity}}.
\newblock In \emph{{Language in Mind: Advances in the Study of Language and Thought}}, pages 157--192. The MIT Press.

\bibitem[{Sood et~al.(2023)Sood, K\"ogel, M\"uller, Thomas, B\^ace, and Bulling}]{Sood_2023_CVPR}
Ekta Sood, Fabian K\"ogel, Philipp M\"uller, Dominike Thomas, Mihai B\^ace, and Andreas Bulling. 2023.
\newblock Multimodal integration of human-like attention in visual question answering.
\newblock In \emph{Proceedings of the IEEE/CVF Conference on Computer Vision and Pattern Recognition (CVPR) Workshops}, pages 2647--2657.

\bibitem[{Sood et~al.(2021)Sood, K{\"o}gel, Strohm, Dhar, and Bulling}]{sood-etal-2021-vqa}
Ekta Sood, Fabian K{\"o}gel, Florian Strohm, Prajit Dhar, and Andreas Bulling. 2021.
\newblock \href {https://doi.org/10.18653/v1/2021.conll-1.3} {{VQA}-{MHUG}: A gaze dataset to study multimodal neural attention in visual question answering}.
\newblock In \emph{Proceedings of the 25th Conference on Computational Natural Language Learning}, pages 27--43, Online. Association for Computational Linguistics.

\bibitem[{Sugano and Bulling(2016)}]{sugano2016seeing}
Yusuke Sugano and Andreas Bulling. 2016.
\newblock \href {http://arxiv.org/abs/1608.05203} {Seeing with humans: Gaze-assisted neural image captioning}.

\bibitem[{Takmaz et~al.(2020)Takmaz, Pezzelle, Beinborn, and Fern{\'a}ndez}]{takmaz-etal-2020-generating}
Ece Takmaz, Sandro Pezzelle, Lisa Beinborn, and Raquel Fern{\'a}ndez. 2020.
\newblock \href {https://doi.org/10.18653/v1/2020.emnlp-main.377} {{G}enerating {I}mage {D}escriptions via {S}equential {C}ross-{M}odal {A}lignment {G}uided by {H}uman {G}aze}.
\newblock In \emph{Proceedings of the 2020 Conference on Empirical Methods in Natural Language Processing (EMNLP)}, pages 4664--4677, Online. Association for Computational Linguistics.

\bibitem[{Tsimpoukelli et~al.(2021)Tsimpoukelli, Menick, Cabi, Eslami, Vinyals, and Hill}]{tsimpoukelli2021multimodal}
Maria Tsimpoukelli, Jacob Menick, Serkan Cabi, S.~M.~Ali Eslami, Oriol Vinyals, and Felix Hill. 2021.
\newblock \href {https://openreview.net/forum?id=WtmMyno9Tq2} {Multimodal few-shot learning with frozen language models}.
\newblock In \emph{Advances in Neural Information Processing Systems}.

\bibitem[{Vaidyanathan et~al.(2018)Vaidyanathan, Prud{'}hommeaux, Pelz, and Alm}]{vaidyanathan-etal-2018-snag}
Preethi Vaidyanathan, Emily~T. Prud{'}hommeaux, Jeff~B. Pelz, and Cecilia~O. Alm. 2018.
\newblock \href {https://doi.org/10.18653/v1/P18-2022} {{SNAG}: Spoken narratives and gaze dataset}.
\newblock In \emph{Proceedings of the 56th Annual Meeting of the Association for Computational Linguistics (Volume 2: Short Papers)}, pages 132--137, Melbourne, Australia. Association for Computational Linguistics.

\bibitem[{van~der Meulen et~al.(2001)van~der Meulen, Meyer, and Levelt}]{Meulen2001EyeMD}
Femke~F. van~der Meulen, Antje~S. Meyer, and Willem J.~M. Levelt. 2001.
\newblock Eye movements during the production of nouns and pronouns.
\newblock \emph{Memory \& Cognition}, 29:512--521.

\bibitem[{van Miltenburg et~al.(2018{\natexlab{a}})van Miltenburg, Elliott, and Vossen}]{van-miltenburg-etal-2018-measuring}
Emiel van Miltenburg, Desmond Elliott, and Piek Vossen. 2018{\natexlab{a}}.
\newblock \href {https://aclanthology.org/C18-1147} {Measuring the diversity of automatic image descriptions}.
\newblock In \emph{Proceedings of the 27th International Conference on Computational Linguistics}, pages 1730--1741, Santa Fe, New Mexico, USA. Association for Computational Linguistics.

\bibitem[{van Miltenburg et~al.(2018{\natexlab{b}})van Miltenburg, K{\'a}d{\'a}r, Koolen, and Krahmer}]{miltenburg2018DIDEC}
Emiel van Miltenburg, {\'A}kos K{\'a}d{\'a}r, Ruud Koolen, and Emiel Krahmer. 2018{\natexlab{b}}.
\newblock \href {http://aclweb.org/anthology/C18-1310} {{DIDEC: The Dutch Image Description and Eye-tracking Corpus}}.
\newblock In \emph{Proceedings of the 27th International Conference on Computational Linguistics (COLING)}, pages 3658--3669. Association for Computational Linguistics.

\bibitem[{van Schijndel and Linzen(2021)}]{vanschijndellinzen}
Marten van Schijndel and Tal Linzen. 2021.
\newblock \href {https://doi.org/https://doi.org/10.1111/cogs.12988} {Single-stage prediction models do not explain the magnitude of syntactic disambiguation difficulty}.
\newblock \emph{Cognitive Science}, 45(6):e12988.

\bibitem[{Wason and Evans(1974)}]{WASON1974141}
P.C. Wason and J.ST.B.T. Evans. 1974.
\newblock \href {https://doi.org/https://doi.org/10.1016/0010-0277(74)90017-1} {Dual processes in reasoning?}
\newblock \emph{Cognition}, 3(2):141--154.

\bibitem[{Yarbus(1967)}]{yarbus1967eye}
Alfred~L. Yarbus. 1967.
\newblock Eye movements during perception of complex objects.
\newblock In \emph{Eye movements and vision}, pages 171--211. Springer.

\bibitem[{Yun et~al.(2013)Yun, Peng, Samaras, Zelinsky, and Berg}]{6618945}
Kiwon Yun, Yifan Peng, Dimitris Samaras, Gregory~J. Zelinsky, and Tamara~L. Berg. 2013.
\newblock \href {https://doi.org/10.1109/CVPR.2013.101} {Studying relationships between human gaze, description, and computer vision}.
\newblock In \emph{2013 IEEE Conference on Computer Vision and Pattern Recognition}, pages 739--746.

\bibitem[{Zhang et~al.(2020)Zhang, Saran, Liu, Zhu, Guo, Niekum, Ballard, and Hayhoe}]{ijcai2020p689}
Ruohan Zhang, Akanksha Saran, Bo~Liu, Yifeng Zhu, Sihang Guo, Scott Niekum, Dana Ballard, and Mary Hayhoe. 2020.
\newblock \href {https://doi.org/10.24963/ijcai.2020/689} {Human gaze assisted artificial intelligence: A review}.
\newblock In \emph{Proceedings of the Twenty-Ninth International Joint Conference on Artificial Intelligence, {IJCAI-20}}, pages 4951--4958. International Joint Conferences on Artificial Intelligence Organization.
\newblock Survey track.

\bibitem[{Zhu et~al.(2018)Zhu, Lu, Zheng, Guo, Zhang, Wang, and Yu}]{selfbleu}
Yaoming Zhu, Sidi Lu, Lei Zheng, Jiaxian Guo, Weinan Zhang, Jun Wang, and Yong Yu. 2018.
\newblock \href {https://api.semanticscholar.org/CorpusID:3636178} {Texygen: A benchmarking platform for text generation models}.
\newblock \emph{The 41st International ACM SIGIR Conference on Research \& Development in Information Retrieval}.

\end{thebibliography}
\bibliographystyle{acl_natbib}

\appendix

\section{Data Preprocessing}
\label{spacy}
We use spaCy to extract the first noun of each description. The numbers of errors in terms of lemmatization and POS-tagging are as follows when using the small, medium, and large spaCy models for Dutch, respectively: 33, 32, 23 mistakes in the full descriptions, and 3, 2, 2 for the first nouns. As the utterances sometimes contain incomplete sentences and disfluencies, POS-tagging may not be reliable in such cases, especially in the later parts of the utterances. However, the large model was reliable both for full descriptions and the first nouns. Hence, we chose to use the data processed by the large model. The model was not able to tag any nouns in 7 descriptions; for those, we use the $<$unk$>$ token as a placeholder starting point. We also skipped nouns such as `photo' (`a photo of a car'), `number' (as in `a number of cats'), `couple' (as in `a couple of kids').

\section{Distribution of Speech Onsets}
\label{distonset}
\begin{figure}[h]
    \centering
    \begin{minipage}[b]{0.45\columnwidth}
    \includegraphics[width=\columnwidth]{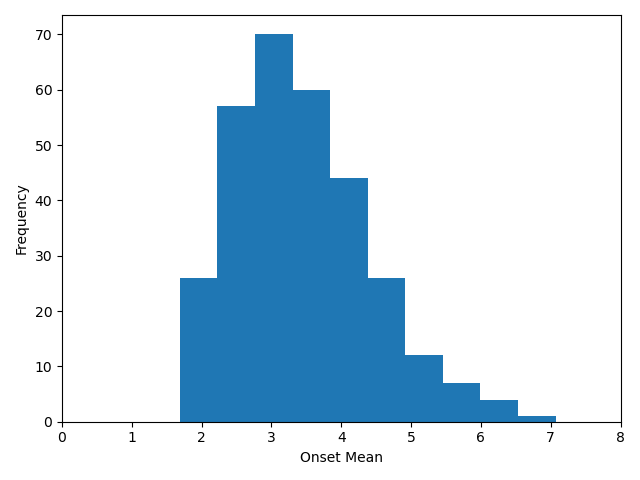}
  \end{minipage}
  \hfill
  \begin{minipage}[b]{0.45\columnwidth}
    \includegraphics[width=\columnwidth]{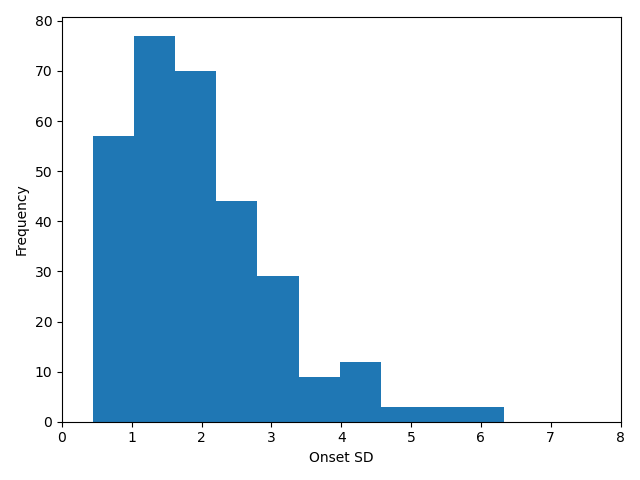}
  \end{minipage}
    \caption{Distributions of onset means and SDs for the images in the whole dataset.}
    \label{fig:hist}
\end{figure}

The histograms of the mean speech onsets and their standard deviations reveal non-normal distributions, as illustrated in Figure~\ref{fig:hist}. 

\section{Participant-Based Correlation Analysis}
\label{ppncorr}
To have a better understanding of speaker-specific dynamics, in addition to calculating statistics per image, we also look into per-participant statistics. Each participant describes around 100 images, each with a possibly different speech onset. We calculate the correlation between a participant’s speech onsets and the BLEU-2-based linguistic variation score of the corresponding images. In 24 out of 45 participants, we find significant moderate negative correlations. All 45 participants have negative correlation coefficients, indicating that all participants tend to start describing an image earlier if that image elicits less linguistic variation across speakers. This suggests that although there can be speaker-specific and contextual factors, the features of an image can also have an overarching effect on the behavioral responses across speakers, and may allow for the prediction of such responses.

\section{BERTje-based Variation in Descriptions}
\label{bertje}

We inspect linguistic variation by comparing the representations of the descriptions extracted using a Dutch BERT model~\cite[BERTje;][]{devries2019bertje}. To calculate variation based on BERTje, we utilize the last hidden state corresponding to the [CLS] token for each description as the representation. Then, for each image, we calculate the pairwise cosine similarities between these representations. The average of these similarities is assigned as the variation found in the descriptions of an image. This method yields scores in the narrow range of $0.69-0.86$, which indicates semantically quite similar descriptions. Since most descriptions have semantics suitable for the corresponding image, the variation in the semantic space is not substantial. Between BERTje-based variation and speech onsets, we reveal a slight negative correlation (Spearman's $\rho = -0.212, p < 0.01$). The SD of speech onsets is even less correlated with BERTje-based variation (Spearman's $\rho = -0.151, p < 0.01$). 

\section{Further Analyses on Linguistic Variation Metrics}
\label{metrics}
We also combine BERTje- and BLEU-2-based variation scores by taking their mean. This metric yields correlations comparable to the ones achieved by the BLEU-2 version, with a moderate increase in the correlation to the starting point variation and mean onset, yet a decrease in the correlation to gaze variation. For the sake of simplicity, we opt for the BLEU-2 version.

We also compare the BLEU-2-based metric against human evaluations for a different dataset provided by \citet{Jas_2015_CVPR}, which achieves a significant correlation (Spearman's $\rho = -0.40, p < .001$), albeit to a moderate extent. \citet{Jas_2015_CVPR} propose a metric that achieves a stronger correlation ($\rho=0.72$). Note that the provided human annotations were obtained through 3 annotators evaluating sentence similarities without looking at the images (comparing only 2 sentences at a time). In our dataset, using our metric, we compare 1 description against 14. As a result, the procedure for human annotations may not be well-aligned with our method (i.e., our metric compares 1 sentence against 4 for their dataset, as each image has 5 descriptions).

\section{Example Images with Scores}
\label{app_examples}
\begin{figure}[h]
    \centering
    \begin{minipage}[h]{0.45\columnwidth}
    \includegraphics[width=\columnwidth]{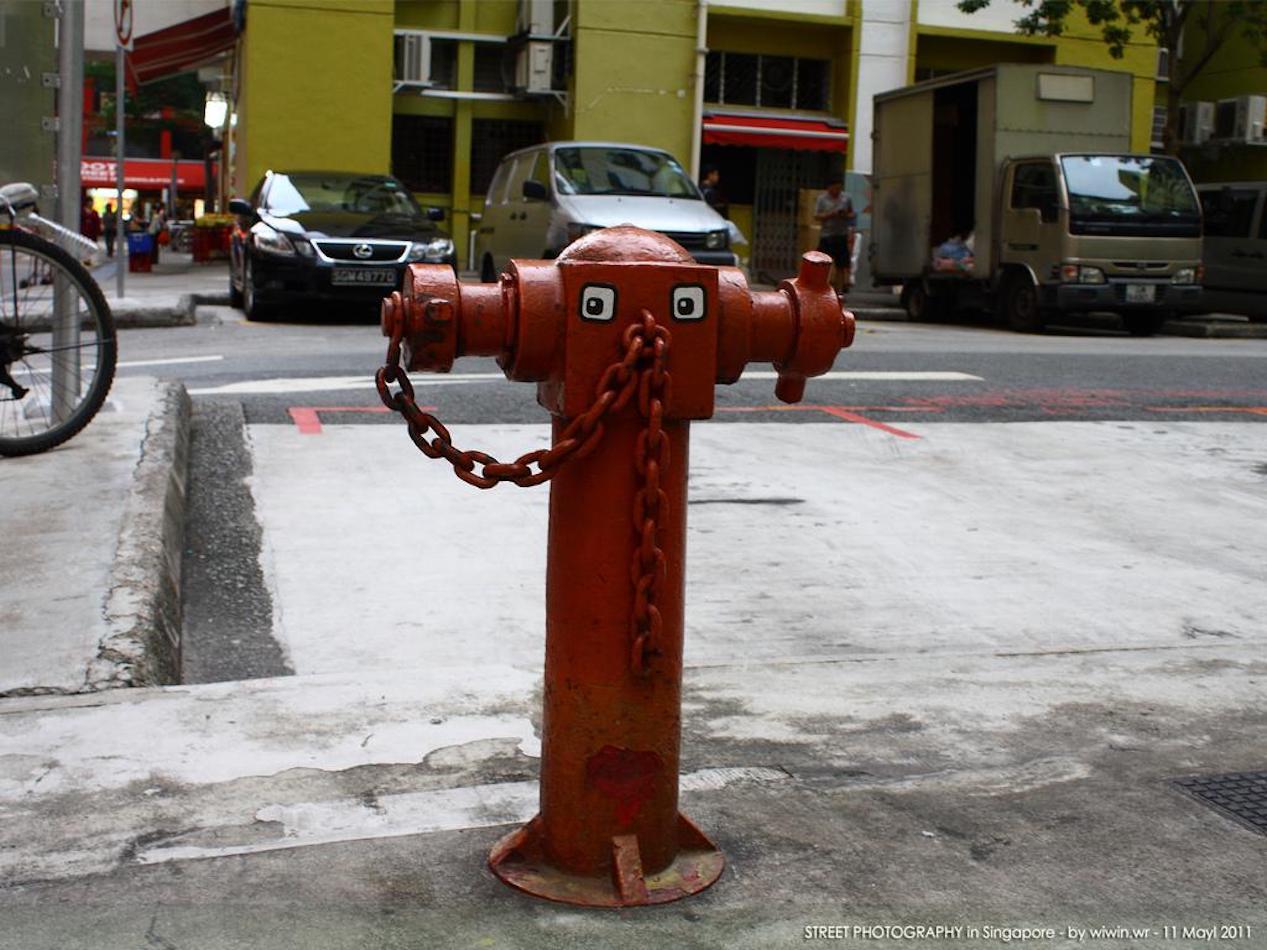}
    \caption*{Min: 0.248}
  \end{minipage}
  \hfill
  \begin{minipage}[h]{0.45\columnwidth}
    \includegraphics[width=\columnwidth]{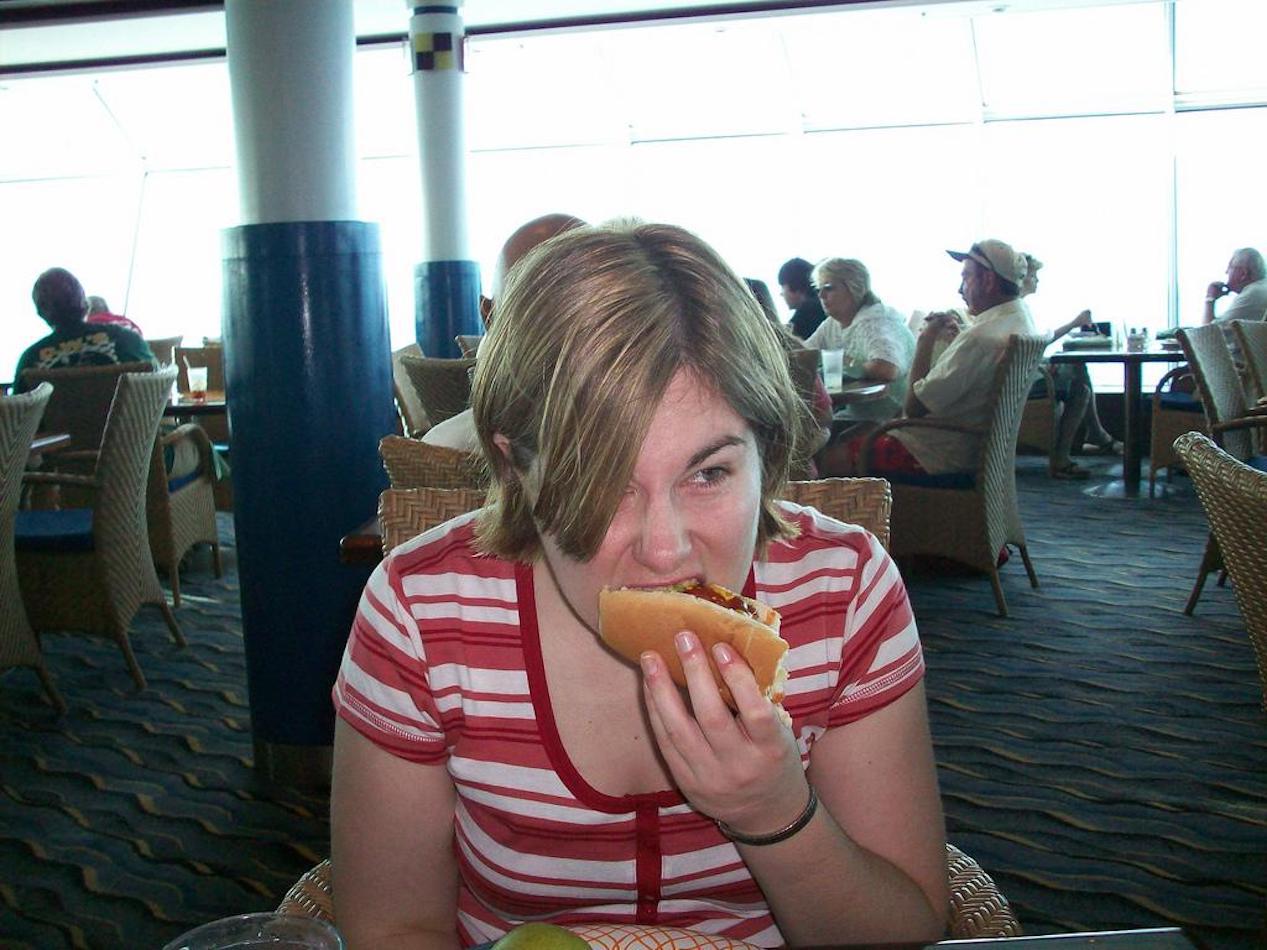}
    \caption*{Max: 0.811}
  \end{minipage}
    \caption{BLEU-2-based linguistic variation scores.}
    \label{fig:exp_imgspec}
\end{figure}

\begin{figure}[h]
    \centering
    \begin{minipage}[h]{0.45\columnwidth}
    \includegraphics[width=\columnwidth]{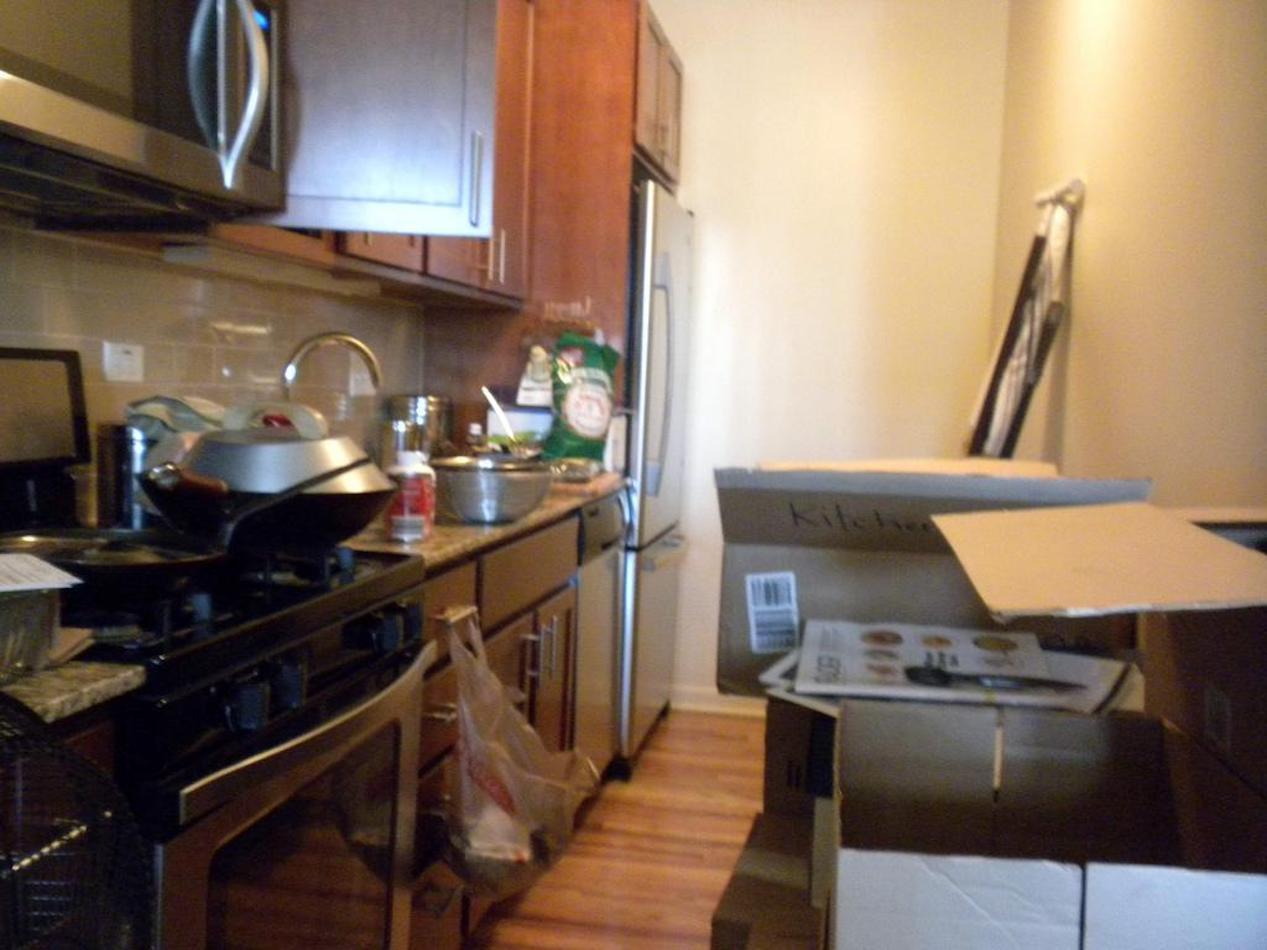}
    \caption*{Min: 1}
  \end{minipage}
  \hfill
  \begin{minipage}[h]{0.45\columnwidth}
    \includegraphics[width=\columnwidth]{images/107930.jpg}
    \caption*{Max: 13}
  \end{minipage}
    \caption{Variation in the number of unique starting points. For the image with the minimum score, all the speakers start with \textit{keuken}, meaning kitchen. The image with the maximum score has descriptions starting with a variety of words: bureau, fitness, huiskamer, springding, atletiek, balk, hoek, tafel, plek, turnattribuut, restaurant, bank, turnobject.}
    \label{fig:exp_spvar}
\end{figure}

\begin{figure}[h]
    \centering
    \begin{minipage}[h]{0.45\columnwidth}
    \includegraphics[width=\columnwidth]{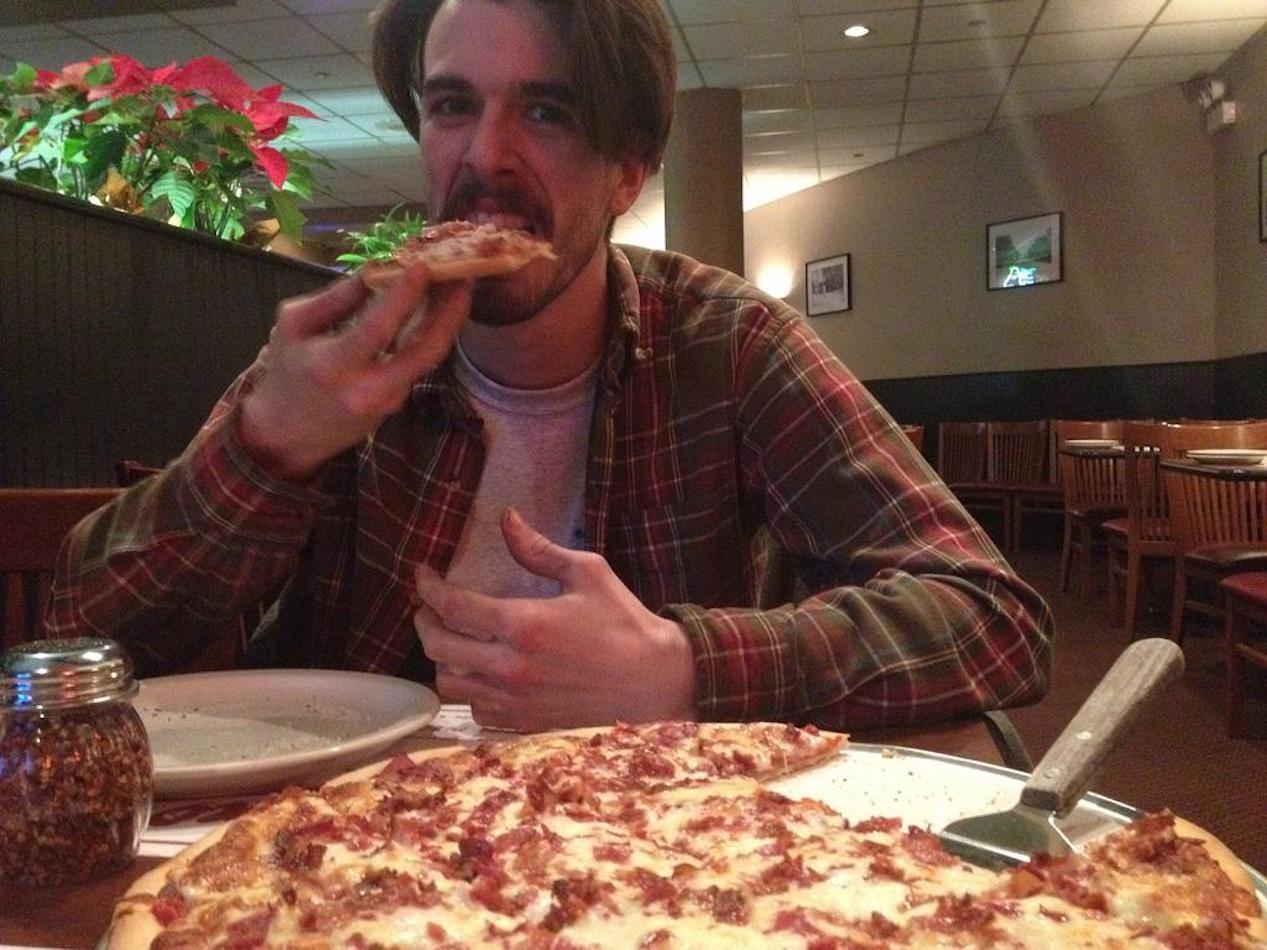}
    \caption*{Min: 11.22}
  \end{minipage}
  \hfill
  \begin{minipage}[h]{0.45\columnwidth}
    \includegraphics[width=\columnwidth]{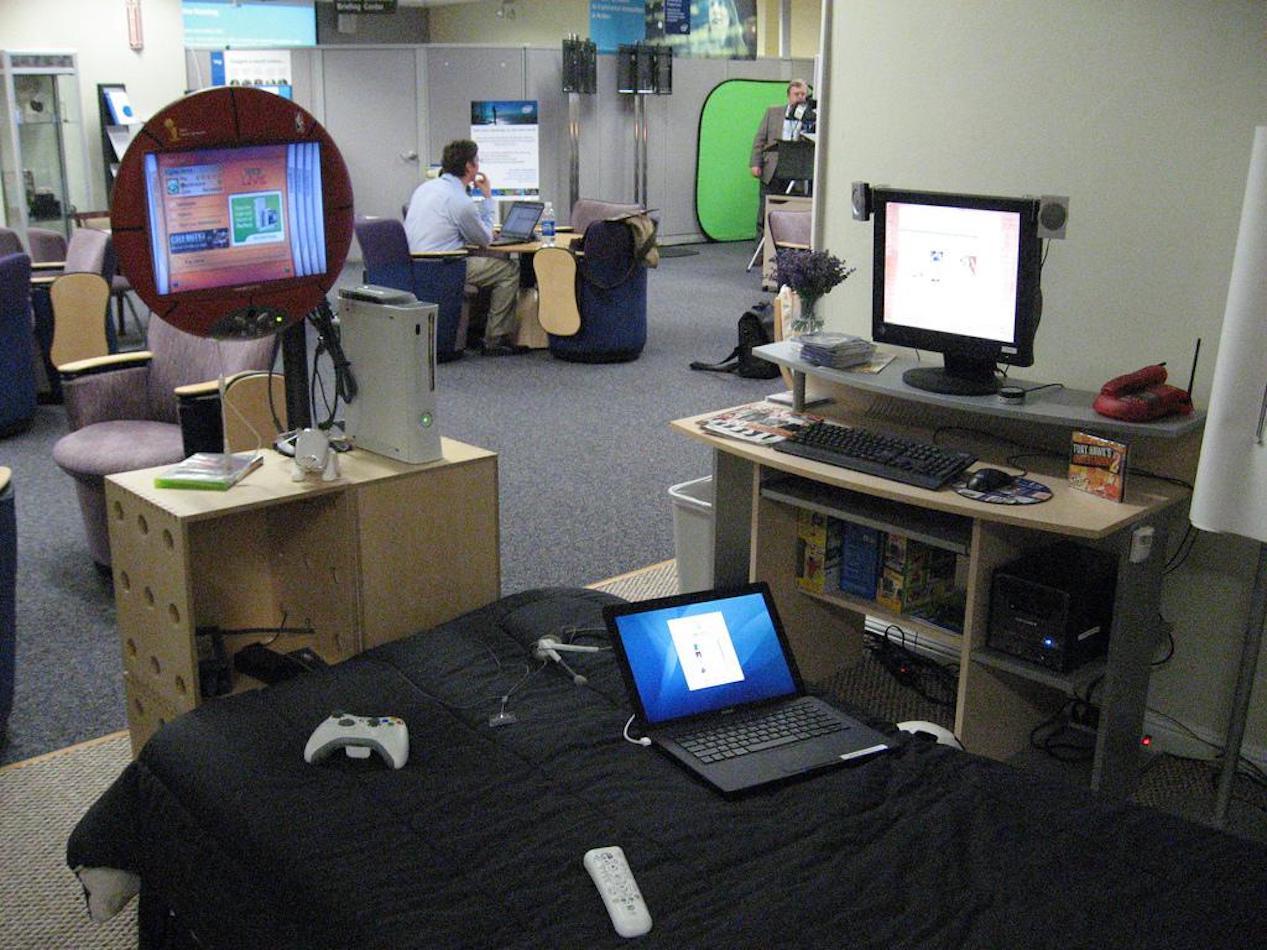}
    \caption*{Max: 38.79}
  \end{minipage}
    \caption{Variation in gaze. The image with the minimum score elicited more similar scanpaths across speakers than the one with the maximum score.}
    \label{fig:exp_gazevar}
\end{figure}

We illustrate the images with the minimum and maximum scores per variable of interest as calculated with our metrics. Figure~\ref{fig:exp_imgspec} depicts variation in the descriptions Figure~\ref{fig:exp_spvar} the variation in starting points, and Figure~\ref{fig:exp_gazevar} the variation in gaze. 

\section{Correlation between Human Signals of Variation}
\label{corrline}
\begin{figure}[h]
    \centering
    \includegraphics[width=0.8\columnwidth]{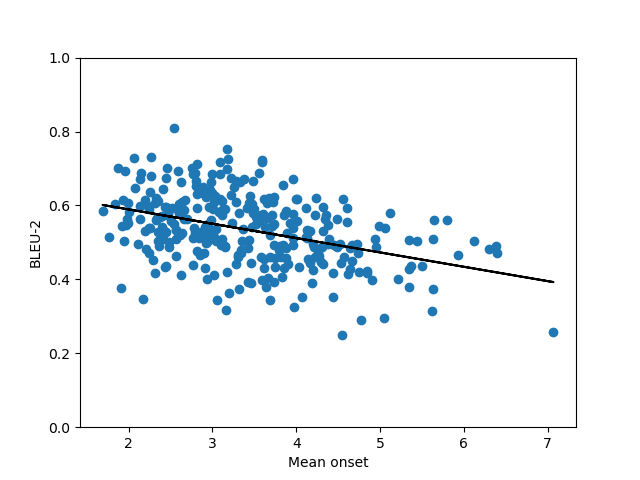}
    \caption{Correlation between mean onset and BLEU-2.}
    \label{fig:corr}
\end{figure}

We illustrate the correlation between the mean onset and the BLEU-2 scores of full descriptions in Figure~\ref{fig:corr}.

\end{document}